\documentclass{article}

\PassOptionsToPackage{numbers, compress}{natbib}

\usepackage[preprint]{neurips_2024}

\usepackage[utf8]{inputenc} 
\usepackage[T1]{fontenc}    
\usepackage[pagebackref=true,breaklinks=true,letterpaper=true,colorlinks,bookmarks=false]{hyperref}
\usepackage{url}            
\usepackage{booktabs}       
\usepackage{amsfonts}       
\usepackage{nicefrac}       
\usepackage{microtype}      
\usepackage{xcolor}         

\usepackage[most]{tcolorbox}
\usepackage{tabularx} 
\usepackage{graphicx} 
\usepackage{geometry} 
\usepackage{enumitem}

\usepackage{wrapfig}
\usepackage{tikz}
\usepackage{comment}
\usepackage{amsmath,amssymb} 
\usepackage{color}
\usepackage{listings}
\usepackage{caption}
\usepackage{adjustbox}
\usepackage{booktabs}
\usepackage{multirow} 
\usepackage{circledsteps} 
\usepackage{csquotes}
\usepackage{soul} 
\usepackage{color, xcolor} 

\usepackage{epigraph}

\definecolor{ForestGreen}{RGB}{0,150,0}

\lstdefinestyle{json}{
    basicstyle=\ttfamily\small,
    showstringspaces=false,
    breaklines=true,
    frame=lines
}
\usepackage[draft]{minted}

\newcommand{\name}{\textsc{\textbf{Hawk}}}

\title{\includegraphics[width=0.06\linewidth]{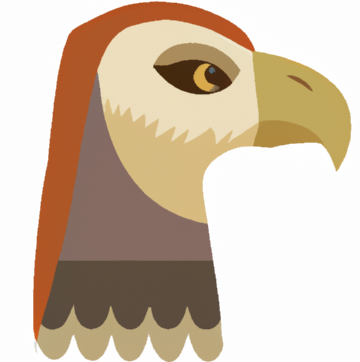} \name:\\Learning to Understand Open-World Video Anomalies}

\author{%
Jiaqi Tang$^{1,2,3}$\footnotemark[1] \quad Hao Lu$^{1,2}$\footnotemark[1] \quad Ruizheng Wu$^{4}$ \quad Xiaogang Xu$^{5,6}$ \quad \textbf{Ke Ma}$^{7}$
\\
\quad \textbf{Cheng Fang}$^{7}$ \quad \textbf{Bin Guo}$^{7}$ \quad \textbf{Jiangbo Lu}$^{3,4}$ \quad \textbf{Qifeng Chen}$^{2}$ \quad \textbf{Ying-Cong Chen}$^{1,2,3}$\footnotemark[2]
\\
$^1$The Hong Kong University of Science and Technology (Guangzhou)
\\ $^2$The Hong Kong University of Science and Technology \quad
$^3$HKUST(GZ) – SmartMore Joint Lab 
\\ $^4$SmartMore Corporation  \quad $^5$The Chinese University of Hong Kong  \quad $^6$Zhejiang University  \\ $^7$Northwestern Polytechnical University 
\\
\small
\texttt{\{jtang092, hlu585\}@connect.hkust-gz.edu.cn} 
\\ \small \texttt{\{ruizheng.wu, jiangbo\}@smartmore.com}  \quad
\texttt{xiaogangxu00@gmail.com}
\\ \small \texttt{\{2544552413,sura\}@mail.nwpu.edu.cn} \quad
\texttt{guob@nwpu.edu.cn}
\\ \small \texttt{cqf@ust.hk} \quad
\texttt{yingcongchen@hkust-gz.edu.cn}
}

\begin{document}

\renewcommand{\thefootnote}{\fnsymbol{footnote}} 
\footnotetext[1]{Equal contribution.} 
\footnotetext[2]{Corresponding author.} 

\maketitle

\begin{abstract}
Video Anomaly Detection (VAD) systems can autonomously monitor and identify disturbances, reducing the need for manual labor and associated costs. However, current VAD systems are often limited by their superficial semantic understanding of scenes and minimal user interaction. Additionally, the prevalent data scarcity in existing datasets restricts their applicability in open-world scenarios.
In this paper, we introduce \name, a novel framework that leverages interactive large Visual Language Models (VLM) to interpret video anomalies precisely. Recognizing the difference in motion information between abnormal and normal videos, \name\ explicitly integrates motion modality to enhance anomaly identification. To reinforce motion attention, we construct an auxiliary consistency loss within the motion and video space, guiding the video branch to focus on the motion modality. Moreover, to improve the interpretation of motion-to-language, we establish a clear supervisory relationship between motion and its linguistic representation. Furthermore, we have annotated over 8,000 anomaly videos with language descriptions, enabling effective training across diverse open-world scenarios, and also created 8,000 question-answering pairs for users' open-world questions. The final results demonstrate that \name\ achieves SOTA performance, surpassing existing baselines in both video description generation and question-answering. Our codes/dataset/demo will be released at \url{https://github.com/jqtangust/hawk}.


\end{abstract}

\section{Introduction}
\label{intro}
\begin{displayquote}
\textit{"Have eyes like a \name!" -- Longman Dictionary}
\end{displayquote}
In recent years, the deployment of Video Anomaly Detection (VAD) systems has seen a significant uptick across a diverse array of domains, including but not limited to, autonomous driving~\cite{yao2022dota,lu2024scaling}, surveillance~\cite{chan2008modeling,lu2013abnormal}, and crime scene analysis~\cite{sultani2018real}. The inherent capability of these systems to autonomously monitor and identify disturbances within a scene has markedly diminished the reliance on manual labor, thereby streamlining operational efficiency and reducing associated costs.

\begin{figure}[t]
\centering
\includegraphics[width=1\linewidth]{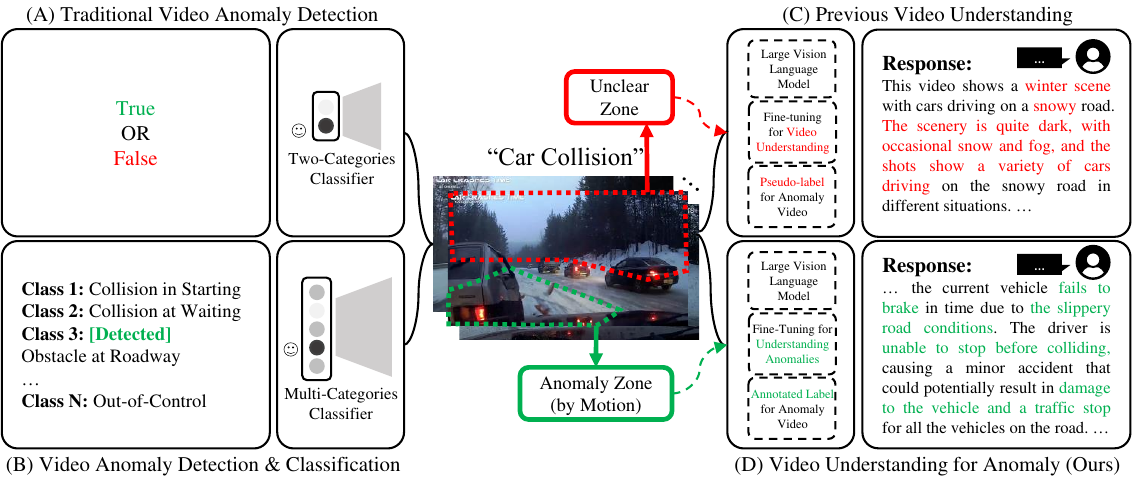}
\caption{Different framework in video anomaly detection. (A) shows traditional video anomaly detection methods, which use binary classifiers to detect anomalies. (B), following (A), introduces a multi-class classifier for integrating semantic information, allowing users to obtain different types of anomaly information. Neither (A) nor (B) can interact with users. (C) is a previous video understanding framework that can interactively provide richer semantic information for users, but cannot specifically locate video anomalies. Our framework (D) enhances the anomaly understanding capability and provides annotated labels with rich semantic information.} 
\label{motivation}
\end{figure}

Despite the extensive focus on anomaly detection in most existing VAD systems~\cite{lu2013abnormal,xu2015learning,sultani2018real,pu2023learning,dubey20193d,he2018anomaly,li2022self,tian2021weakly,wu2021learning,zaheer2020claws,zhu2019motion} (as shown in Fig.~\ref{motivation} (A)), there is often a lack of deeper semantic understanding of the scenes and insufficient interaction with users. While Pu et al.~\cite{pu2023learning} and Wu et al.~\cite{wu2023openvocabulary} incorporated semantic information for video anomaly detection, their frameworks are limited as multiple-class classifiers (as displayed in Fig.~\ref{motivation} (B)). Consequently, the functionality of these systems is confined to the detection of anomalous frames, necessitating further manual analysis by users to analyze the detected anomalies comprehensively.
Although Lv et al.~\cite{lv2024video} has pioneered the development of a large language model for the video anomaly explanation, their approach primarily relies on \textit{pseudo labels} for training. The lack of robust training data severely constrains its practical applicability. Besides, such a method focuses more on acquiring long-range context information rather than anomaly-related features on anomaly understanding (as exhibited in Fig.~\ref{motivation} (C)).

To solve the above challenges, we propose an interactive large visual-language model~\cite{liu2024visual,li2023videochat,Maaz2023VideoChatGPT}, \name, for precisely understanding video anomalies (as illustrated in Fig.~\ref{motivation} (D)).
Considering that the motion in normal and abnormal videos is significantly different~\cite{xu2015learning,zhu2019motion}, we explicitly integrate motion modality by a dual-branch framework in \name\ to enhance the understanding of anomalies (Section~\ref{intergrate}).
Besides, to reinforce motion attention, we construct an auxiliary consistency loss based on the mutual information between the original video (appearance feature) and its motion in tight space (Section~\ref{Consistency}), to implicitly guide the video branch to focus on motion-related features.
However, the interpretation of motion to the corresponding language remains unclear.
Therefore, we extract the motion-related language (verbs and their entities) from the original description to directly supervise the visual and linguistic representations of motion, for accurately enhancing the interpretation of video anomaly in \name\ (Section~\ref{langloss}).

Furthermore, we also collect seven video anomaly datasets from various scenarios and generate language descriptions for each video. Besides, to address the open-ended questions raised by users, we utilize language descriptions of the videos to generate potential question-answer pairs for training.
Since these datasets cover a range of scenarios (Section~\ref{descrip}), including crime (UCF-Crime~\cite{sultani2018real}), campus environments (ShanghaiTech~\cite{liu2018ano_pred} and CUHK Avenue~\cite{lu2013abnormal}), pedestrian walkways (UCSD Ped1~\cite{chan2008modeling} and Ped2~\cite{wang2010anomaly}), traffic situations (DoTA~\cite{yao2022dota}), and human behavior (UBnormal~\cite{acsintoae2022ubnormal}), and finally, the model tends to generalize to open-world scenarios. 

To train our framework, we initially pre-train it on WebVid~\cite{Bain21} to equip it with the capability to understand general videos. Then, we fine-tuned it on our proposed video anomaly dataset to enhance its understanding of video anomalies across multiple scenarios. Compared to other baselines, our model achieves SOTA performance in both Text-Level and GPT-Guided Metrics. Our contributions are summarized as follows:
\begin{itemize}[itemsep=0pt,topsep=0pt,parsep=0pt]
    \item We propose a novel video-language framework, \name, aiming at understanding video anomalies, which incorporates motion modality to enhance its capability.
    \item We generate rich language descriptions for seven different video anomaly datasets. Meanwhile, considering the diversity of open-world problems, we also generate question-answer pairs to tackle potential user inquiries.
    \item Compared to other large video models, our framework demonstrates SOTA performance for video anomaly understanding and question-answering across multiple scenarios, which will help open-world anomaly understanding in the future.
\end{itemize}

\section{Related Work}
\label{related}
\paragraph{Video Anomaly Detection}
Video Anomaly Detection (VAD) usually focuses on identifying unexpected events from the video
and it has been widely applied in various fields, including autonomous driving~\cite{yao2022dota}, public surveillance~\cite{chan2008modeling,lu2013abnormal}, and crime scene analysis~\cite{sultani2018real} etc. Previous VAD methods~\cite{lv2024video,sultani2018real,lu2013abnormal,xu2015learning,dubey20193d,he2018anomaly,li2022self,tian2021weakly,wu2021learning,zaheer2020claws,zhu2019motion} are designed in numerous pathways. Lu et al.~\cite{lu2013abnormal} designed to learn video features only from normal videos, and hand-craft features or deep-learning-based features are leveraged. Sultani et al.~\cite{sultani2018real} proposed multiple instance learning (MIL), which is the main paradigm for many weakly-supervised learning methods. Recently, Lv et al.~\cite{lv2024video} first proposed video-based large language models in the framework of VAD. 

However, these methods lack sufficient semantic comprehension of scenes and offer inadequate user interaction. Several approaches~\cite{pu2023learning,wu2023openvocabulary} have introduced multi-class classifiers to integrate semantic information with various types of anomaly information. Nevertheless, their output is still limited.
In contrast, our framework not only integrates more comprehensive semantic information as a general video understanding system but also provides advanced interaction capabilities for users.

\paragraph{Large Model in Video Understanding}
Recent studies have demonstrated the reliable capabilities of large models in video understanding. Beyond powerful vision-language models~\cite{li2023blip,zhu2023minigpt,liu2024visual,lu2024gpt}, recent research has increasingly explored more modalities~\cite{lv2024video,li2023videochat,maaz2023video,ye2023mplug,luo2023valley}. Bain et al.\cite{Bain21} introduced a large-scale dataset with general video content descriptions. Several LLM-based works\cite{li2023videochat,maaz2023video,ye2023mplug,luo2023valley} aim to comprehend visual content. Additionally, Video-LLaMa~\cite{damonlpsg2023videollama} extends comprehension to both auditory and visual information, while Su et al.\cite{su2023pandagpt} utilize multi-modal encoders to understand across six modalities. Recently, Lv et al.\cite{lv2024video} proposed video-based large language models for VAD tasks in a weakly supervised framework.
In this paper, we introduce the \textit{motion modality} in our proposed vision-language model, which enhances the model's ability to locate anomalies by prioritizing relevant video content.

\section{Data Engineering}
\label{data}
Previous datasets are inadequate for addressing our problem. Most existing VAD datasets, such as UBnormal~\cite{acsintoae2022ubnormal} and DoTA~\cite{yao2022dota}, only contain \textbf{simple video category labels} and lack detailed language descriptions. This results in video understanding models lacking accurate and comprehensive supervision, creating a significant obstacle to identifying anomalies in videos.
Recently, Lv et al.\cite{lv2024video} attempted to create \textbf{pseudo language descriptions} for anomaly videos. However, these descriptions are naive combinations of labels and fixed text, relying on a rigid format that offers only limited information. 
Other datasets, like WebVid\cite{Bain21}, include only \textbf{general descriptions} of video content, which may not direct the model's focus on anomalies.

\begin{figure}[!t]
\centering
\includegraphics[width=1\linewidth]{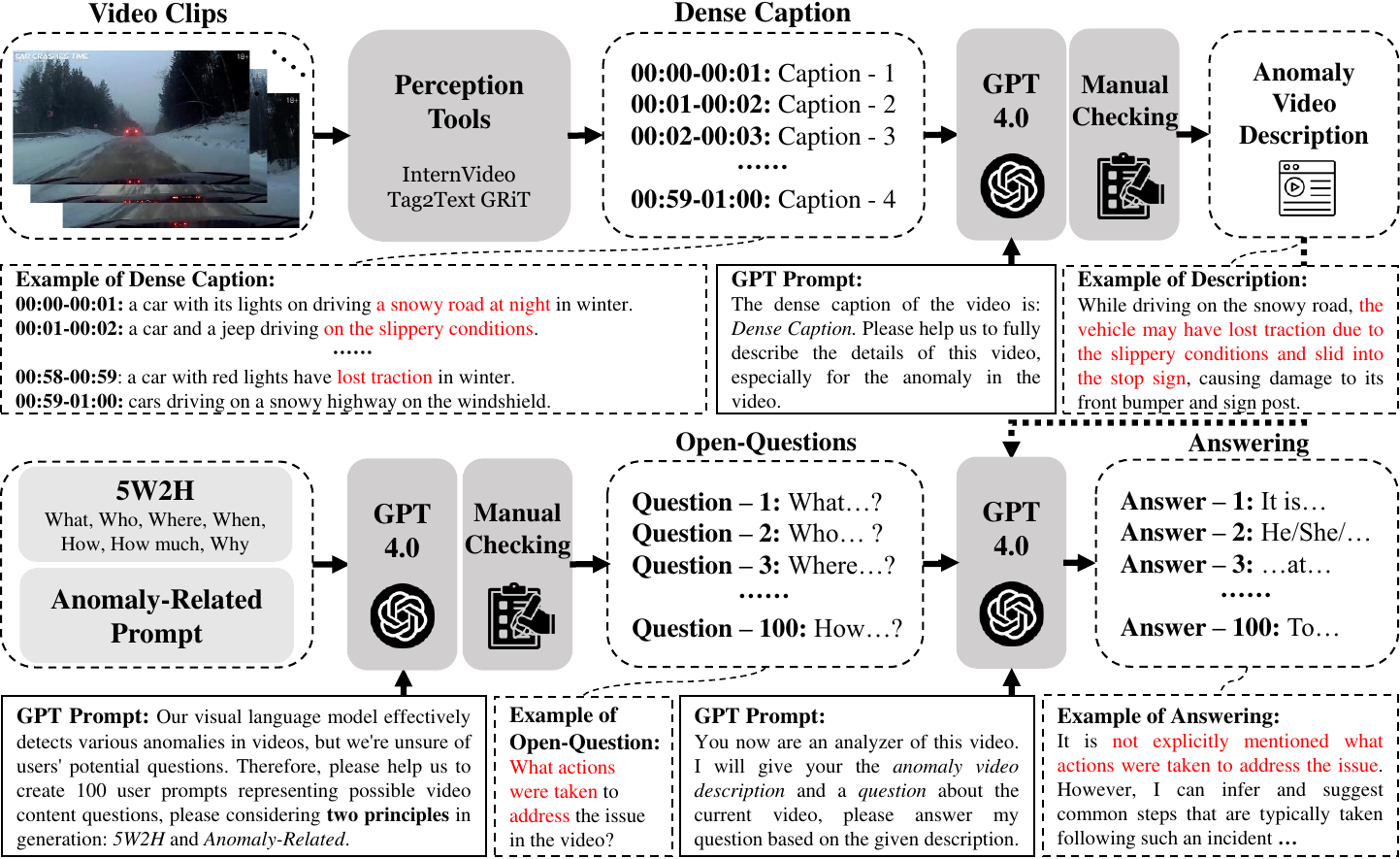}
\caption{Generation pipeline of our dataset. In the first line, we first \textbf{segment videos into clips} and \textbf{generate dense captions} for each segment, including a comprehensive description of the video content. 
Then, we use GPT-4 to guide the \textbf{generation of corresponding anomalous video descriptions} based on these descriptions, which are then \textbf{manually checked to reduce mistakes}. In the second line, to generate user-centered QA pairs, we first use GPT-4 to \textbf{generate open-ended questions} based on the proposed two principles. Then, the questions and video descriptions are jointly input into GPT-4 to \textbf{provide possible answers}.} 
\label{datagen}
\end{figure}

\paragraph{Our Principle} To tackle the above problems, we annotate detailed language descriptions specifically for anomaly scenes in seven different existing \textsc{<Video>} datasets.
These seven datasets include a variety of anomalous scenarios such as crime (UCF-Cirme~\cite{sultani2018real}), campus (ShanghaiTech~\cite{liu2018ano_pred} and CUHK Avenue~\cite{lu2013abnormal}), pedestrian walkways (UCSD Ped1~\cite{chan2008modeling} and Ped2~\cite{wang2010anomaly}), traffic (DoTA~\cite{yao2022dota}), and human behavior (UBnormal~\cite{acsintoae2022ubnormal}).
With the support of these visual scenarios, we can perform comprehensive fine-tuning for various abnormal scenarios, being closer to open-world scenarios.

Moreover, to better account for real-world user situations, we believe that language descriptions should not only include \textbf{descriptions of the video anomalies} themselves, but also address \textbf{open questions asked by users}. Therefore, we construct open-ended question-answer pairs for each scenario to further enhance model's practical ability to answer users' varying questions. The procedure for answering users' questions is shown in Fig.~\ref{datagen}. The data format of can be described by the Eq.~\eqref{data},
\begin{equation}\label{data} \small
\textsc{<Video>: \ \{\textbf{Dis}: \ <Description> \ | \ \textbf{QA}: \ <{Question}> } \rightarrow \textsc{<Answering>\}}.
\end{equation}

\paragraph{Anomaly Video Description Generation} \label{descrip}
To construct natural language descriptions $\textsc{<Description>}$ for anomalous video datasets, we refer to previous research such as LLaVa~\cite{liu2024visual} and VideoChat~\cite{li2023videochat}, and employ GPT-4~\cite{achiam2023gpt} as an assistant. We first split the video into dense clips to ensure key information is captured. Following VideoChat~\cite{li2023videochat}, we use perception tools (InternVideo~\cite{wang2022internvideo}, Tag2Text~\cite{huang2023tag2text}, or GRiT~\cite{wu2022grit}) to automatically generate captions for each key clip, obtaining a dense representation of the videos (except for the UCF-Crime dataset, which already has a dense representation built in \cite{yuan2023surveillance}).
Next, we use GPT-4~\cite{achiam2023gpt} to generate anomaly-related descriptions based on the captions for each video. Unlike other general video understanding datasets~\cite{liu2024visual,li2023videochat}, we provide prompts for GPT-4 to generate specific descriptions closely related to video anomalies. Finally, due to varying quality of dense captions, some videos may have incorrect annotations. Thus, we manually recheck the final generated video anomaly descriptions to ensure label accuracy.

\paragraph{Human-Centric Question-Answering Pairs Generation} \label{open}
So far, we have obtained nearly accurate descriptions of anomaly videos. However, our framework may still face challenges with more open-ended questions from users. Therefore, anomaly-related question-answering is a significant practical requirement. Given the diversity of open-world scenes, users may ask questions involving various pronouns.
Thus, we mainly consider these two principles: \textbf{\CircledText{1} Anomaly-related}, our questions should be strongly related to the anomaly in the video. \textbf{\CircledText{2} 5W2H}, 
we introduce seven different question pronouns (What, Who, Where, When, How, How much, and Why) to simulate various question formats that users may employ. This enables us to address a wide range of open questions related to video anomalies.
We input these two principles into GPT-4~\cite{achiam2023gpt} to generate open questions for anomaly videos. We then manually review and select the 100 most suitable questions, which are randomly assigned to each video.
Finally, GPT-4~\cite{achiam2023gpt} will generate $\textsc{<Answers>}$ to these $\textsc{<Questions>}$.

Our data is more practical compared to previous ones: it not only understands multiple anomalies in videos but also supports question-answering in open scenarios (More details in Appendix~\ref{appdata}).

\section{Methodology}
\label{method}
To construct a practical framework for understanding video anomalies, our goal is to accurately interpret these anomalies into natural language. However, most previous studies~\cite{li2023videochat,damonlpsg2023videollama,Maaz2023VideoChatGPT,lin2023video,lv2024video} focus on enhancing general video understanding capabilities while neglecting video anomalies. This oversight results in equal attention being given to all parts of the video, such as the background and human appearances, often at the expense of key anomaly features, as shown in Fig.~\ref{motivation} (C). Consequently, these approaches are not effective in accurately focusing on anomaly-related features.

\paragraph{Overview of Solution} 
The core of our solution is guiding visual instruction to focus on anomalies. 
Previous studies in video anomaly detection~\cite{xu2015learning,zhu2019motion} have demonstrated that \textit{motion-related feature} help identify multiple anomalies. Therefore, in Section~\ref{intergrate}, we first explicitly integrate a motion modality into our proposed framework to target anomaly-related features. Subsequently, in Section~\ref{Consistency}, we maintain mutual information consistency between the appearance and motion modalities within a tight feature space, implicitly guiding the appearance branch to reinforce motion attention.
Finally, in Section~\ref{langloss}, to improve the interpretation of motion-to-language, we extract motion-related language descriptions to directly match the motion and its corresponding motion-related language.

\subsection{Explicit Motion Modality Integration} \label{intergrate}
\begin{figure}[!t]
\centering
\includegraphics[width=1\linewidth]{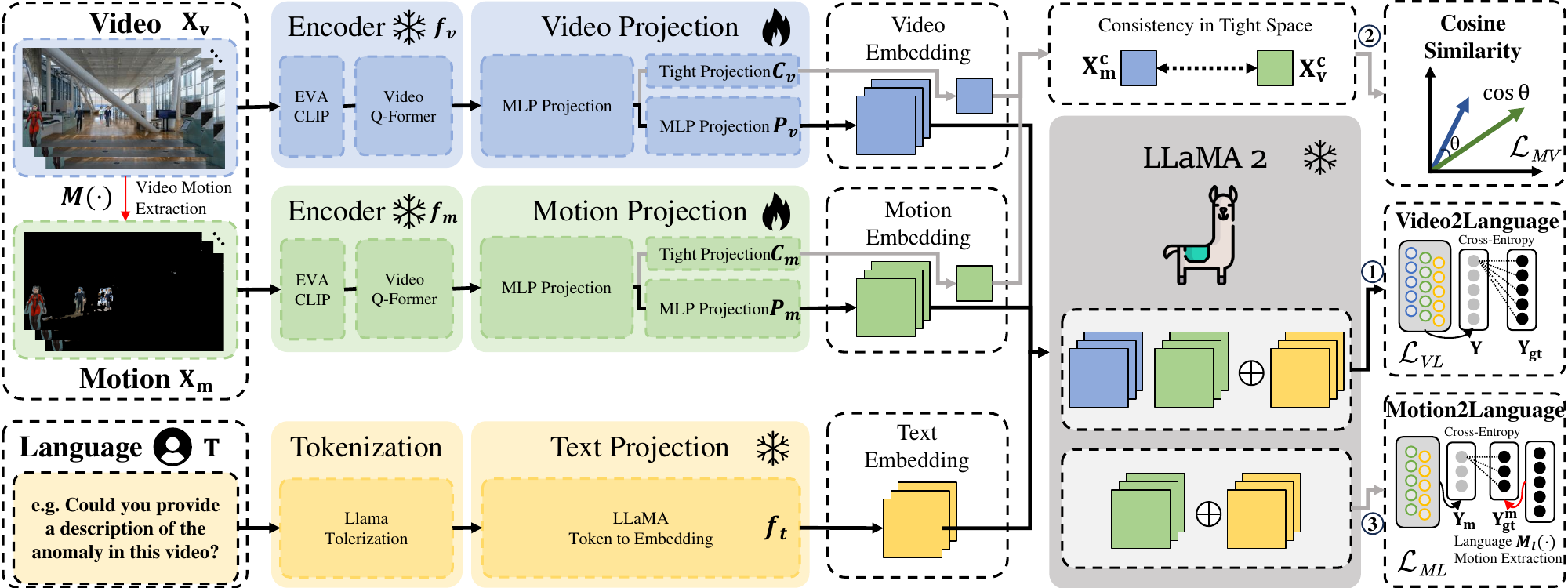}
\caption{Overview of \name. During training (\textcolor{black}{\textbf{Black}} and \textcolor{gray}{\textbf{Gray}} path), we aim to optimize for video-language matching loss, along with Video-Motion Consistency and Motion-Language Matching. 
During inference (only \textcolor{gray}{\textbf{Gray}} path), we generate language descriptions using video, motion, and text.}
\label{framework}
\end{figure}

To enhance the capability of interpreting anomalies, we build a framework, \name, to explicitly integrate motion modality.
\name\ has a dual-branch architecture, with $f_v$ as the original video understanding network and $f_m$ for motion understanding.
Inspired by Video-LLaMA~\cite{damonlpsg2023videollama}, $f_v$ and $f_m$ share the same architecture but separate parameters in Fig.~\ref{framework}.
Eq.~\eqref{model} denotes our framework as,
\begin{equation} \label{model} \small
\mathbf{Y} = \textsc{LLaMa}\left([P_v(f_v(\mathbf{X_v})), P_m(f_m(\mathbf{X_m}))] \oplus f_t(\mathbf{T})\right),
\end{equation}
where $\mathbf{X_v} \in \mathrm{R}^{T \times C \times H \times W}$ represents the $\textsc{<Video>}$ input for extracting appearance feature, and $T$ denotes the temporal dimension. $\mathbf{X_m} = M(\mathbf{X_v})$, with $M(\cdot)$ being the motion extractor.

$f_v(\cdot)$ and $f_m(\cdot)$ are the frozen pre-trained video encoders from BLIP-2~\cite{Li2023BLIP2BL}, which consist of one EVA-CLIP~\cite{Fang2022EVAET} and one pre-trained Video Q-Former to output embeddings.
Then, the output embeddings from $f_v(\cdot)$ and $f_m(\cdot)$ are passed through learnable projection networks for video and motion, $P_v(\cdot)$ and $P_m(\cdot)$, respectively.
These networks aim to project visual (video and motion) embedding into the language feature space for interpreting.
$f_t(\cdot)$ is the frozen text token to embedding projection, that makes textual information can be inputted into LLaMA-2~\cite{touvron2023llama}.
$\oplus$ is for combining our input prompt, we define our prompt as: ``\textit{Here is the input video embedding: \textsc{<Video\_Embedding>} and motion embedding \textsc{<Motion\_Embedding>} in different frames, please help me to \textsc{<Describe\_Video> | <Question>}}.''. \textsc{<Describe\_Video>} and \textsc{<Question>} are the question classes for video description generation and video question answering respectively (Details see Appendix~\ref{prompt}).
By combining the visual token embedding with the textual embedding, $f_t(\textbf{T})$, LLaMA-2~\cite{touvron2023llama}, is employed to generate the final language response, $\mathbf{Y}$.
This framework explicitly integrates the motion modality during visual instruction tuning, significantly targeting anomaly-related features.

\subsection{Implicitly Motion Attention Reinforcement} \label{Consistency}
\begin{wrapfigure}{r}{0.4\textwidth}
  \centering
  \includegraphics[width=1\linewidth]{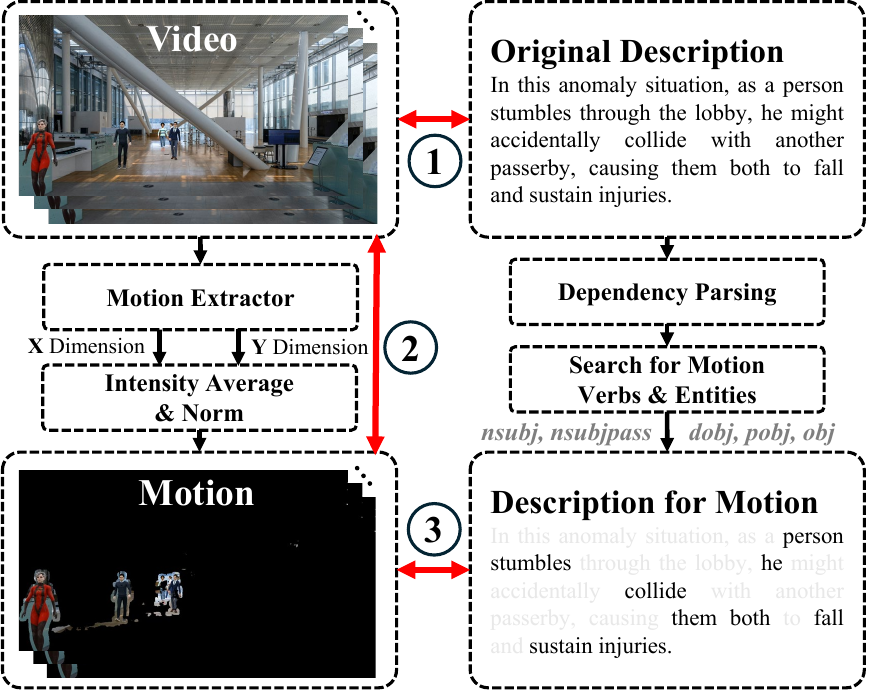}
  \caption{Visualization of \name's loss. \CircledText{1} is the original video-to-language loss. \CircledText{2} is the cosine similarity loss for motion modality adaptation. \CircledText{3} is the motion-to-language loss.}
  \label{vloss}
\end{wrapfigure}

Although we integrate the motion modality to facilitate fine-tuning of \name, motion and video branches operate independently. 
Therefore, we cannot expect the original video branch to extract appearance features that focus on the region where the anomaly occurred (i.e., motion).
To help \name\ focus more on these regions, we observed the containment relationship in mutual information between motion and the original video. We use this relationship to construct an auxiliary consistency loss function, implicitly reinforcing the motion attention (Fig.~\ref{vloss} \CircledText{2}).
\paragraph{Extract Motion} Specifically, to obtain motion, we employ a motion describer $M(\cdot)$, which generates motion between two successive frames as shown in Eq.~\eqref{motion},
\begin{equation} \label{motion}  \small
    \mathbf{X_{Motion}^{(t)}} = M^{\mathbf{(t)}}(\mathbf{X_v^{(t)}}, \mathbf{X_v^{(t-1)}}),
\end{equation}
where $M^{\mathbf{(t)}}(\cdot)$ is the motion describer at the time step $\mathbf{t}$, we currently use Gunnar Farneback's algorithm~\cite{farneback2000fast}, and $ \small \mathbf{X_v^{(t)}}, \mathbf{X_v^{(t-1)}} \in \mathrm{R}^{1 \times C \times H \times W}$ denote the video frames at time steps $\mathbf{t}$ and $\mathbf{t-1}$.

$\small \mathbf{X_{Motion}^{(t)}} \in \mathbb{R}^{2 \times H \times W}$ includes two channels motion vector in $\mathbf{X}$ (horizontal) and $\mathbf{Y}$ (vertical) directions. We use the optical flow magnitude from these channels as a $\mathbf{Mask}$, normalized to $[0,1]$ and multiplied with the original video appearance, to hide other non-motion regions, as Eq.~\eqref{e_motion},
\begin{equation} \small
\begin{aligned}
    \mathbf{X_m^{(t)}} = \underset{\mathbf{Mask}}{\underbrace{\textsc{Norm}( \sqrt{(\mathbf{X_{Motion}^{(t)}(X)})^2 + (\mathbf{X_{Motion}^{(t)}(Y)})^2})}} \times \mathbf{X_v^{(t)}},    
\end{aligned} \label{e_motion}
\end{equation}
where $\times$ is the operator of pixel-wise multiplication. $\small \mathbf{X_v^{(t)}}, \mathbf{X_m^{(t)}} \in \mathrm{R}^{1 \times C \times H \times W}$ donate the original video and our input motion information at time step $\mathbf{t}$, respectively. We usually extract $T$ frames as motion input $\small \mathbf{X_m} \in \mathrm{R}^{T \times C \times H \times W}$, same as $\mathbf{X_v}$.
\paragraph{Build $\mathcal{L}_{MV}$ Loss} 
Then, we consider that $\mathbf{X_m}$ only contains key information for anomaly and it is contained in $\mathbf{X_v}$, and feature space from $\mathbf{X_v}$ is more sparse.
Therefore, we compact features from $\mathbf{X_m}$ and $\mathbf{X_v}$ into a tight space. At this space, we aim to maintain the mutual information between $\mathbf{X_m}$ and $\mathbf{X_v}$ consistency, and in this way, the appearance feature can be focused on the motion region.
Therefore, we construct an auxiliary loss to promote $\mathbf{X_v}$'s motion attention, as in Eq.~\eqref{consine},
\begin{equation}
\small
\mathcal{L}_{MV} =1 - \textsc{Cos\_Sim}(\mathbf{X_m^c}, \mathbf{X_v^c})
= 1 - \frac{\mathbf{X_m^c} \cdot \mathbf{X_v^c}}{\|\mathbf{X_m^c}\| \|\mathbf{X_v^c}\|},
 \label{consine}
    \end{equation}
where $\small\mathbf{X_v^c} = C_v(f_v(\mathbf{X_v}))$ and $\small\mathbf{X_m^c} = C_m(f_m(\mathbf{X_m}))$ denote the tightly compressed representations of $\mathbf{X_v}$ and $\mathbf{X_m}$, respectively, by the compression functions $C_v$ and $C_m$. $C_v$ and $C_m$ share some initial shallow layer parameters with $P_v$ and $P_m$ (as Fig.~\ref{framework}). Then, following a subsequent tight projection to compresses both $\mathbf{X_v}$ and $\mathbf{X_m}$ into a more compacted space.

Finally, with this auxiliary loss, we can reinforce the motion attention in the appearance feature, and \name's feature space will focus on more abnormal related features, which will promote the understanding of anomalies in the whole framework.
\subsection{Interpreting Motion-to-Language} \label{langloss}
Although \name\ has already accommodated the motion modality in visual input, the corresponding motion from language is still unclear. This limitation hinders \name's interpretation in motion modality. Hence, to augment this relationship, we aim to reinforce the correspondence between motion and their linguistic representation.

\paragraph{Extract Motion-related Language} Previous studies~\cite{cadiot2006semantics,vo2020identifying,wunderlich1997cause,langacker1987nouns} have proved that the representation of motion in the language is predominantly from \textbf{verbs} and their corresponding \textbf{entities}.
Therefore, to extract linguistic representation, the first step is to do dependency parsing for the original sentences, as Eq.~\eqref{depen},
\begin{equation} \label{depen} \small
\mathbf{G_{gt}} = D(\mathbf{Y_{gt}}),
\end{equation}
where $D(\cdot)$ is the dependency parsing and $\mathbf{Y_{gt}}$ is the ground truth. $\mathbf{G_{gt}}$ represents the graph of the dependency structure, which symbolizes the syntactic relationships among the words in a sentence.

Based on this graph, we can extract predicates (verbs) $\mathbf{V}$, and also entities closely related to these predicates, such as subjects $\mathbf{S}$, objects $\mathbf{O}$, indirect subjects $\mathbf{S_i}$, and indirect objects $\mathbf{O_i}$. These elements are then combined to form short phrases representing motion, as in Eq.~\eqref{extract},
\begin{equation}\label{extract} \small
 \mathbf{Y_{gt}^m} = \{\mathbf{V},\mathbf{S},\mathbf{O},\mathbf{S_i},\mathbf{O_i}\} = M_l(\mathbf{G_{gt}}),
\end{equation}
where $M_l(\cdot)$ is the language motion extraction operator, and $\mathbf{Y_{gt}^m}$ is the motion-related language.
\paragraph{Build $\mathcal{L}_{ML}$ Loss} After obtaining motion-related language, we can establish strong supervision between motion in both vision and linguistic representation (as Fig.~\ref{vloss} \CircledText{3}), significantly enhancing the ability to interpret motion to language in \name.
Consequently, we design a motion-language matching as an auxiliary loss, as Eq.~\eqref{motionloss},
\begin{equation} \small
\begin{aligned}
\mathcal{L}_{ML}^m(\mathbf{Y_m}, \mathbf{Y_{gt}^m}) &= -\sum_{i=1}^{N}\mathbf{Y_{gt}^m}^{i}\log(\mathbf{Y_m}^{i})\\
\textrm{s.t.} \ \mathbf{Y_m} &= \textsc{LLaMa}\left( P_m(f_m(\mathbf{X_m})), f_t(\mathbf{T})\right),
\end{aligned}\label{motionloss}
\end{equation}
where $\mathcal{L}_{ML}(\cdot)$ is the cross-entropy loss, which contains $N$ words.
\paragraph{Optimization Goal} Finally, our total loss $\mathcal{L}$ shows as, $\label{totalloss} \small \mathcal{L} = \mathbf{t_0} \times \mathcal{L}_{VL} + \mathbf{t_1} \times \mathcal{L}_{MV} + \mathbf{t_2} \times \mathcal{L}_{ML}$, where $\mathcal{L}_{VL}$ is original video to language loss (as Fig.~\ref{vloss} \CircledText{1}), and $\mathbf{t_0}$, $\mathbf{t_1}$ and $\mathbf{t_2}$ is the hyper-parameter.

\section{Experiments}
This section introduces training, testing, baselines, evaluations, and ablation experiments of \name.
\label{ex}
\paragraph{Training \& Testing}
\setlength{\intextsep}{0mm} 
\begin{wrapfigure}{o}{0.36\textwidth}
  \centering
  \includegraphics[width=1\linewidth]{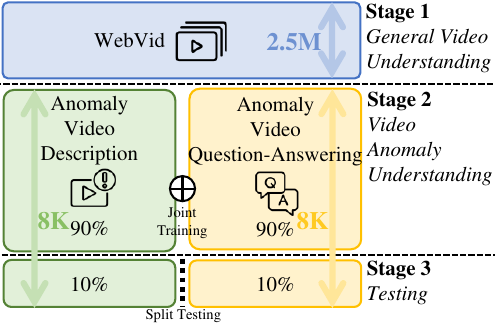}
  \caption{Training \& Testing.}
  \label{procedure}
\end{wrapfigure}
To enhance our framework's anomaly understanding capabilities, we've structured our training and testing process into three stages, as Fig.~\ref{procedure}. \textbf{Stage 1} involves pre-training on the WebVid dataset~\cite{Bain21} to acquire a general understanding of video content. In \textbf{Stage 2}, we finetune the model's focus towards video anomaly understanding by employing a specially curated dataset described in Section~\ref{data}, consisting of over $8,000$ videos. We use 90\% of these videos for training and allocate the remaining 10\% for testing purposes. We jointly train on two tasks: video \textsc{<Description>} generation and video \textsc{<Question>$\rightarrow$<Answering>}. In \textbf{Stage 3}, we evaluate these two tasks independently in the testing set to ensure our model's effectiveness.

\paragraph{Baselines}
To evaluate the anomaly understanding performance of our proposed framework, we conduct comparisons with SOTA video understanding baselines. We select five baselines: Video-ChatGPT~\cite{Maaz2023VideoChatGPT}, VideoChat~\cite{li2023videochat}, Video-LLaMA~\cite{damonlpsg2023videollama}, LLaMA-Adapter~\cite{zhang2023llamaadapter}, and Video-LLaVA~\cite{lin2023video}. 
The purpose of our comparison is to determine whether these baselines can fully understand and interpret video anomalies.
\paragraph{Evaluation Metrics}
To accurately evaluate our model's performance in understanding video anomalies, we firstly adopt four \textbf{Text-Level} metrics, from BLEU (Bilingual Evaluation Understudy)~\cite{papineni2002bleu}-1 to BLEU-4 to measure word overlap between the model-generated text and the ground truth. This approach enables us to objectively assess the similarity and also take into account various levels of granularity at the text-level, thus providing a clear indicator of how well the model understands and describes anomalies.
 
Besides, we expand our evaluation framework by incorporating insights from recent research in LLaVa~\cite{liu2024visual} or Video-ChatGPT~\cite{Maaz2023VideoChatGPT}, utilizing \textbf{GPT-Guided}~\cite{achiam2023gpt} methods to assess the quality of the generated text. GPT~\cite{achiam2023gpt} serves as a critical evaluator, generating scores for three key aspects of the language produced, with each aspect scored on a scale from $0$ to $1$. These three aspects are as,
\begin{itemize}[itemsep=0pt,topsep=0pt,parsep=0pt]
    \item \textbf{Reasonability:} evaluates the logical reasoning and coherence of the generated language.
    \item \textbf{Detail:} assesses the level of detail and specificity of the generated language.
    \item \textbf{Consistency:} evaluates the coherence and consistency of the generated language.
\end{itemize}
By leveraging GPT~\cite{achiam2023gpt} as an evaluative tool, we aim to provide a nuanced understanding of the text's quality, focusing on aspects that traditional metrics may overlook.
\begin{table}[t]
  \centering
  \scriptsize
    \caption{Quantitative performance of (A) anomaly video description generation and (B) video question-answering. \textcolor{red}{Red} indicates the best performance, while \textcolor{blue}{blue} denotes the second best.}
    \newcommand\widthface{1}
    \resizebox{1.0\linewidth}{!}{
    \begin{tabular}{l|cccc|ccc}
    \multicolumn{8}{c}{(A) Anomaly Video Description Generation} \\
    \toprule
    \multirow{2}[4]{*}{Method} & \multicolumn{4}{c|}{Text-Level $(\mathbf{\uparrow})$~\cite{papineni2002bleu}} & \multicolumn{3}{c}{GPT-Guided $(\mathbf{\uparrow})$~\cite{liu2024visual}} \\
\cmidrule{2-8}          & BLEU-1 & BLEU-2 & BLEU-3 & BLEU-4 & Reasonability & Detail & Consistency \\
    \midrule
    Video-ChatGPT~\cite{Maaz2023VideoChatGPT} & 0.107 & 0.046 & 0.017 & \textcolor{blue}{0.008} & 0.084 & 0.108 & 0.055 \\
    VideoChat~\cite{li2023videochat} & 0.053 & 0.023 & 0.008 & 0.003 & 0.107 & 0.205 & 0.054 \\
    Video-LLaMA~\cite{damonlpsg2023videollama} & 0.062 & 0.025 & 0.009 & 0.004 & \textcolor{blue}{0.120}  & \textcolor{blue}{0.217} & \textcolor{blue}{0.066} \\
    LLaMA-Adapter~\cite{zhang2023llamaadapter} & \textcolor{blue}{0.132} & \textcolor{blue}{0.052} & \textcolor{blue}{0.018} & \textcolor{blue}{0.008} & 0.060  & 0.091 & 0.038 \\
    Video-LLaVA~\cite{lin2023video} & 0.071 & 0.030  & 0.012 & 0.005 & 0.077 & 0.115 & 0.038 \\
    \midrule
    Ours  & \textcolor{red}{0.270}  & \textcolor{red}{0.139} & \textcolor{red}{0.074} & \textcolor{red}{0.043} & \textcolor{red}{0.283} & \textcolor{red}{0.320} & \textcolor{red}{0.218} \\
    \bottomrule
    \multicolumn{8}{c}{(B) Anomaly Video Question-Answering} \\
    \toprule
    \multirow{2}[4]{*}{Method} & \multicolumn{4}{c|}{Text-Level $(\mathbf{\uparrow})$~\cite{papineni2002bleu}} & \multicolumn{3}{c}{GPT-Guided $(\mathbf{\uparrow})$~\cite{liu2024visual}} \\
\cmidrule{2-8}          & BLEU-1 & BLEU-2 & BLEU-3 & BLEU-4 & Reasonability & Detail & Consistency \\
    \midrule
    Video-ChatGPT~\cite{Maaz2023VideoChatGPT} & 0.177 & 0.096 & 0.058 & 0.038 & 0.508 & 0.430  & 0.421 \\
    VideoChat~\cite{li2023videochat} & \textcolor{blue}{0.261} & \textcolor{blue}{0.133} & \textcolor{blue}{0.074} & \textcolor{blue}{0.043} & \textcolor{blue}{0.699} & \textcolor{blue}{0.631} & \textcolor{blue}{0.598} \\
    Video-LLaMA~\cite{damonlpsg2023videollama} & 0.156 & 0.081 & 0.045 & 0.027 & 0.586 & 0.485 & 0.497 \\
    LLaMA-Adapter~\cite{zhang2023llamaadapter} & 0.199 & 0.109 & 0.067 & \textcolor{blue}{0.043} & 0.646 & 0.559 & 0.549 \\
    Video-LLaVA~\cite{lin2023video} & 0.094 & 0.054 & 0.034 & 0.023 & 0.393 & 0.274 & 0.316 \\
    \midrule
    Ours  & \textcolor{red}{0.319} & \textcolor{red}{0.179} & \textcolor{red}{0.112} & \textcolor{red}{0.073} & \textcolor{red}{0.840} & \textcolor{red}{0.794} & \textcolor{red}{0.753}  \\
    \bottomrule
    \end{tabular}}
    \label{result_question}%
\end{table}%

\begin{table}[!t]
\centering
\begin{minipage}{\textwidth}
\scriptsize
\caption{Qualitative performance on (A) anomaly video description generation, and (B) question-answering. \textcolor{red}{Red} texts indicate key semantic inconsistencies, whereas \textcolor{ForestGreen}{Green} texts signify that the generated results are closely aligned with the Ground Truth. \sethlcolor{yellow}\hl{[YELLOW]} indicates the text problem.}
\label{QE}
\begin{tcolorbox}[title=(A) Anomaly Video Description Generation,
    colback=white, 
    colframe=black, 
    colbacktitle=black, 
    coltitle=white,
    fonttitle=\scriptsize\bfseries,
    enhanced,
    boxed title style={size=small,colback=green!20!white}] 
    \centering
    \includegraphics[width=0.95\linewidth]{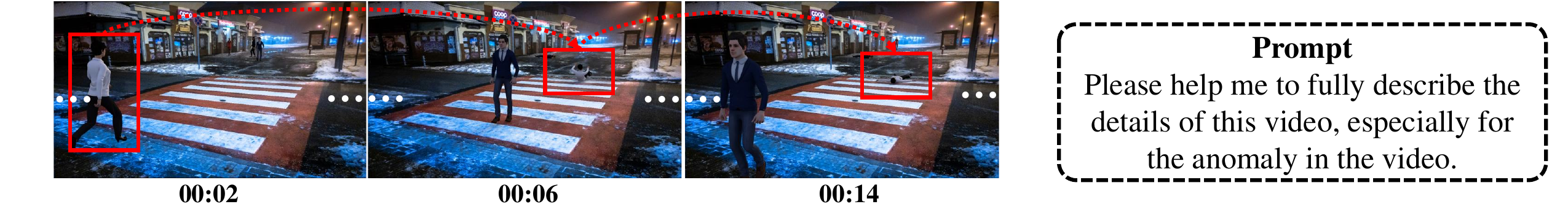} 
    \begin{tabularx}{\textwidth}{@{}lX@{}}
    \toprule
    \textbf{Video-ChatGPT~\cite{Maaz2023VideoChatGPT}:} & \sethlcolor{yellow}\hl{[NOT FOCUS ON ANOMALY]} \textcolor{red}{The most interesting part of the video is a person walking on the snow-covered sidewalk}, which is not a common sight on a snowy street. \\
    \midrule
    \textbf{VideoChat~\cite{li2023videochat}:} & The video captures a pedestrian crossing in a snowy town at night. The crossing features a white and red crossing sign, and there are some street lights on the road. \sethlcolor{yellow}\hl{[NOT FOCUS ON ANOMALY]} \textcolor{red}{A few people are walking on the sidewalk}, while others are crossing the street. \\
    \midrule
    \textbf{Video-LLaMA~\cite{damonlpsg2023videollama}:} & First, the video shows a city street with cars parked on both sides. There are also people walking on the sidewalk. Next, the view changes to \sethlcolor{yellow}\hl{[FOCUS ON PEOPLE'S CLOTHING]} \textcolor{red}{a man and a woman standing on a crosswalk, wearing blue and red jackets respectively}.\\
    \midrule
    \textbf{LLaMA-Adapter~\cite{zhang2023llamaadapter}:} & There is an anomaly in the video: a person appears to be walking on the street, but they are not visible in the blurred image. This person seems to be walking in the midst of the snowstorm, \sethlcolor{yellow}\hl{[HALLUCINATION]} \textcolor{red}{which is unusual as most people would be staying indoors during such weather conditions}. \\
    \midrule
    \textbf{Video-LLaVA~\cite{lin2023video}:} & The video shows a man walking across a street while a woman is lying on the ground. \sethlcolor{yellow}\hl{[UNCLEAR]} \textcolor{red}{The woman lying on the ground} is \textcolor{ForestGreen}{not moving}, and it is unclear what happened to her.\\
    \midrule
    \textbf{Ours:} & In this video, \textcolor{ForestGreen}{the man and woman crossing the street at a crosswalk might be impacted by the anomaly}. \textcolor{ForestGreen}{The woman could slip on the snowy sidewalk and be injured}, which would be the anomaly in the video as they should be mindful of their surroundings. \\
    \midrule
    \textbf{Ground Truth:} & In this anomaly situation, \textcolor{ForestGreen}{one of the people walking across the snowy crosswalk suddenly falls down} probably due to a medical emergency, possibly suffering an injury. This could also lead to a chain reaction, with other pedestrians trying to avoid the fallen person and potentially \textcolor{ForestGreen}{slipping themselves}. \\
    \bottomrule
    \end{tabularx}
\end{tcolorbox}
\begin{tcolorbox}[title=(B) Anomaly Video Question-Answering,
    colback=white, 
    colframe=black, 
    colbacktitle=black, 
    coltitle=white,
    fonttitle=\scriptsize\bfseries,
    enhanced,
    boxed title style={size=small,colback=yellow!20!white}] 
    \centering
    \includegraphics[width=0.95\linewidth]{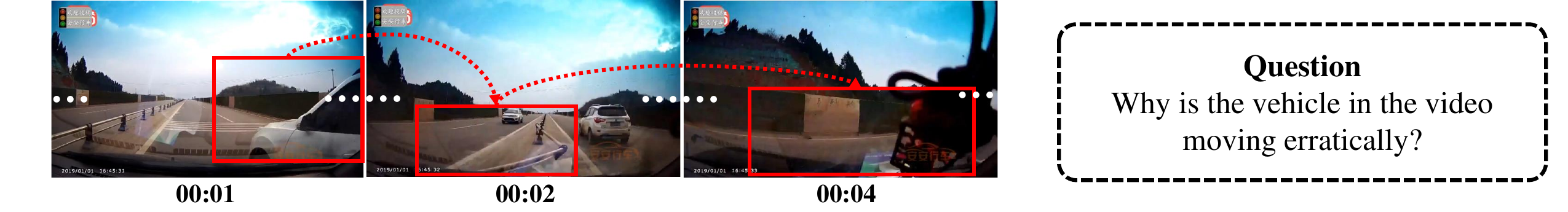} 
    \begin{tabularx}{\textwidth}{@{}lX@{}}
    \toprule
    \textbf{Video-ChatGPT~\cite{Maaz2023VideoChatGPT}:} & The other people affected by the anomaly in the video are the \sethlcolor{yellow}\hl{[HALLUCINATION]} \textcolor{red}{two people in the car who were hit by the car}. They are seen lying on the ground, and the car is seen driving away. \\
    \midrule
    \textbf{VideoChat~\cite{li2023videochat}:} & There are \sethlcolor{yellow}\hl{[WRONG DESCRIPTION]} \textcolor{red}{no other people or vehicles shown in the video}. The only thing that is shown in the video is the black car and the street and trees on either side of it. \\
    \midrule
    \textbf{Video-LLaMA~\cite{damonlpsg2023videollama}:} & Since the video is not providing any additional context, it is not possible to determine if there are other people affected by the anomaly. It could be \sethlcolor{yellow}\hl{[NO IDENTIFYING]} \textcolor{red}{a minor problem that only affects the car}, or it could be something more significant that affects other drivers or pedestrians on the road.\\
    \midrule
    \textbf{LLaMA-Adapter~\cite{zhang2023llamaadapter}:} & \sethlcolor{yellow}\hl{[WRONG]} The other people affected by the anomaly are the \textcolor{red}{driver of the car and the motorcyclist}. \\
    \midrule
    \textbf{Video-LLaVA~\cite{lin2023video}:} &  \sethlcolor{yellow}\hl{[HALLUCINATION]} \textcolor{red}{The driver of the car is seen driving with the hood open}.\\
    \midrule
    \textbf{Ours:} & The vehicle in the video is moving erratically likely due to the driver \textcolor{ForestGreen}{losing control}. The speed and steering of the vehicle may have been \textcolor{ForestGreen}{affected by the uneven road conditions, causing the car to swerve unpredictably and almost hitting the white car}. The abrupt maneuvers might be indicative of \textcolor{ForestGreen}{the driver's attempts to regain control or avoid collisions} with other vehicles or obstacles on the road. \\
    \midrule
    \textbf{Ground Truth:} & Based on the description you've provided, it seems the erratic movement of the vehicle in the video may be a consequence of \textcolor{ForestGreen}{the white car suddenly braking on an empty highway}. \textcolor{ForestGreen}{This abrupt action could have led to the driver of the vehicle behind reacting in a panicked or abrupt manner to avoid a collision, resulting in erratic movements such as swerving or harsh braking}.\\
    \bottomrule
    \end{tabularx}
\end{tcolorbox}
\end{minipage}
\end{table}

\begin{table}[!t]
  \centering
  \scriptsize
  \caption{Ablation study of (A) anomaly video description generation and (B) video question-answering. \textcolor{red}{Red} indicates the best performance, while \textcolor{blue}{blue} denotes the second best.}
    \newcommand\widthface{1}
    \resizebox{1.0\linewidth}{!}{
    \begin{tabular}{l|cccc|ccc}
    \multicolumn{8}{c}{(A) Anomaly Video Description Generation} \\
    \toprule
    \multirow{2}[4]{*}{} & \multicolumn{4}{c|}{Text-Level $(\mathbf{\uparrow})$~\cite{papineni2002bleu}} & \multicolumn{3}{c}{GPT-Guieded $(\mathbf{\uparrow})$~\cite{liu2024visual}} \\
\cmidrule{2-8}          & BLEU-1 & BLEU-2 & BLEU-3 & BLEU-4 & Reasonability & Detail & \multicolumn{1}{c}{Consistency} \\
    \midrule
    w/o Motion Information& \textcolor{blue}{0.249} & 0.121 & 0.062 & 0.034 &  0.253     &  \textcolor{blue}{0.306}     & 0.189 \\
    w/o Video-Motion Consistency & \textcolor{blue}{0.249} & 0.123 & 0.064 & 0.036 &     0.261   & 0.295      &  0.194\\
    w/o Motion-Language Matching Loss & \textcolor{red}{0.270}  & \textcolor{blue}{0.135} & \textcolor{blue}{0.073} & \textcolor{blue}{0.041} &  \textcolor{blue}{0.276 }    &  \textcolor{red}{0.320}     & \textcolor{blue}{0.212} \\
    \midrule
    Ours  & \textcolor{red}{0.270}  & \textcolor{red}{0.139} & \textcolor{red}{0.074} & \textcolor{red}{0.043} & \textcolor{red}{0.283} & \textcolor{red}{0.320} & \textcolor{red}{0.218} \\
    \bottomrule
    \multicolumn{8}{c}{(B) Anomaly Video Question-Answering} \\
    \toprule
    \multirow{2}[4]{*}{} & \multicolumn{4}{c|}{Text-Level $(\mathbf{\uparrow})$~\cite{papineni2002bleu}} & \multicolumn{3}{c}{GPT-Guieded $(\mathbf{\uparrow})$~\cite{liu2024visual}} \\
\cmidrule{2-8}          & BLEU-1 & BLEU-2 & BLEU-3 & BLEU-4 & Reasonability & Detail & \multicolumn{1}{c}{Consistency} \\
    \midrule
    w/o Motion Information & 0.309 & 0.171 & 0.105 & 0.065 &   \textcolor{blue}{0.837}    &    \textcolor{blue}{0.790}   & 0.743 \\
    w/o Video-Motion Consistency & 0.313 & 0.172 & 0.105 & 0.066 & 0.833      & 0.784       & 0.742  \\
    w/o Motion-Language Matching Loss & \textcolor{blue}{0.316} & \textcolor{blue}{0.176} & \textcolor{blue}{0.109} & \textcolor{blue}{0.069} &   0.836    & 0.788      & \textcolor{blue}{0.752} \\
    \midrule
    Ours  & \textcolor{red}{0.319} & \textcolor{red}{0.179} & \textcolor{red}{0.112} & \textcolor{red}{0.073} & \textcolor{red}{0.840} & \textcolor{red}{0.794} & \textcolor{red}{0.753} \\
    \bottomrule
    \end{tabular}}
  \label{ab}%
\end{table}%

\paragraph{Quantitative Evaluation}
Table~\ref{result_question} (A) and (B) demonstrate the effectiveness of our model to describe abnormal phenomena. Our proposed model significantly outperforms the previous baselines, achieving SOTA performance in every metric for both Text-level and GPT-guided metrics, thus it can generate text that more closely aligns with actual scenarios.
\paragraph{Qualitative Evaluation}
Table~\ref{QE} (A) and (B) demonstrate that our proposed framework achieves optimal qualitative performance in video description generation and question-answering, respectively. 
Compared with other baselines, \name\ can accurately understand and focus on video anomalies. 
For example, in Table~\ref{QE} (A) - Video-LLaMa~\cite{damonlpsg2023videollama}, it pays more attention to the clothing information from the people (\textcolor{red}{\textit{wearing blue and red jacket}}), while ignoring the motion-related anomaly (\textcolor{ForestGreen}{\textit{slipping}}). In Table~\ref{QE} (B) - Video-ChatGPT, it may produce hallucinations (\textcolor{red}{\textit{two people... who were hit by the car}}), which differ from the original video anomaly (\textcolor{ForestGreen}{\textit{car suddenly braking}}). In contrast, \name\ generates descriptions that are close to the real semantics (\textcolor{ForestGreen}{\textit{driver losing control}}).
\paragraph{Ablation Study} We conducted ablation experiments on three key structures proposed in this paper and analyzed their impact on the overall performance in Table~\ref{ab} (A) and (B).
\begin{table}[!t]
\centering
\scriptsize
\caption{Qualitative evaluation of ablation study. \textcolor{red}{Red}/\textcolor{ForestGreen}{Green} texts indicate language semantic inconsistency/consistency with the Ground Truth respectively. \sethlcolor{yellow}\hl{[YELLOW]} indicates the text problem.}
\newcommand\widthface{1}
\resizebox{1.0\linewidth}{!}{
    
    \begin{tabularx}{\textwidth}{@{}lX@{}}
    \toprule
    \textbf{w/o Motion Information:} & While driving down the \sethlcolor{yellow}\hl{[UNRELATED INFORMATION]} \textcolor{red}{snowy street}, \textcolor{red}{the vehicle suddenly skids on a patch of ice }, causing it to veer into another car that was \sethlcolor{yellow}\hl{[HALLUCINATION]} \textcolor{red}{parked beside a tree} on the side of the road. \\
    \midrule
    \textbf{w/o Video-Motion Consistency:} & In this video, the car fails to stop in time due to the \sethlcolor{yellow}\hl{[WRONG CONDITION]} \textcolor{red}{slippery road caused by the snowy conditions}. As a result, it \textcolor{ForestGreen}{skids and sideswipes the white car} parked on the side of the road, causing minor damage to both vehicles. \\
    \midrule
    \textbf{w/o Motion-Language Matching:} & As the white truck drives down the street, it suddenly skids on the wet road surface, losing control, and  \textcolor{ForestGreen}{sideswiping the parked white car}. The impact results in visible damage to both vehicles, \sethlcolor{yellow}\hl{[UNCLEAR CONDITION]} \textcolor{red}{with smoke emitting from the truck's side and the car's mirrors shattering.}\\
    \midrule
    \textbf{Ours:} & While driving down a narrow street with cars parked on both sides, the current vehicle's front right side \textcolor{ForestGreen}{scrapes against a parked car}, \textcolor{ForestGreen}{causing minor damage to both vehicles}. \\
    \midrule
    \textbf{Ground Truth:} & While driving down the street, the silver car suddenly swerves to \textcolor{ForestGreen}{avoid a parked car}, but clips its rear bumper, \textcolor{ForestGreen}{causing minor damage to both vehicles}.\\
    \bottomrule
    \end{tabularx}}
\label{Ablation_QE}
\end{table}

\begin{itemize}[itemsep=0pt,topsep=0pt,parsep=0pt]
    \item \textbf{Effectiveness of Motion Information:} We ablate all the motion components, including $f_m$, $P_m$ and the motion input $\mathbf{X_m}$ for proving the effectiveness of introducing motion modality.
    When explicit motion information is lacking, the model's ability to describe the motions-related anomaly diminishes, leading to inaccurate descriptions or even hallucinations (Table~\ref{Ablation_QE} w/o Motion Information), then impedes the overall performance (Table~\ref{ab}).
    \item \textbf{Effectiveness of Video-Motion Consistency:} The absence of video-motion consistency constraints reduces the generative model's ability to adapt to the motion modality, causing difficulties in accurately understanding motion scenes (Table~\ref{Ablation_QE} w/o Video-Motion Consistency), then impedes the overall performance (Table~\ref{ab}).
    \item \textbf{Effectiveness of Motion-Language Matching:} Without motion-language matching loss, the correlation between motion and language becomes unclear. This ambiguity leads to the generation of language that includes unspecified motion information (Table~\ref{Ablation_QE} w/o Motion-Language Matching), subsequently degrading the overall performance (Table~\ref{ab}).
\end{itemize}

\section{Conclusion}
\label{con}
In conclusion, we have developed a novel video-language framework for understanding video anomalies across various scenarios. By incorporating motion features and constructing rich linguistic descriptions, our model demonstrates SOTA performance in the open world. It has the potential to benefit practical applications in diverse domains and paves the way for improving the model's interactivity with users, enabling more efficient and effective communication in addressing user-specific inquiries related to video anomalies.





{
\bibliographystyle{splncs04}
\bibliography{egbib}



}

\newpage

\appendix
\section{Summary of Appendix} This appendix provides supplementary information that was not included in the main paper. Firstly, we address the security statement of our study, ensuring the confidentiality and integrity of the data used. Additionally, we provide detailed explanations of the training and testing resources utilized, including information on the hardware and software configurations. We also present statistics and distribution of the training data, along with the costs associated with human resources involved in the study. Furthermore, we describe the evaluation metrics employed to assess the performance of our method.
Moreover, we present additional qualitative results comparisons, showcasing the effectiveness of our approach. Additionally, we provide an open-world demo, demonstrating the real-world applicability of our method.
Finally, we discuss the existing limitations of our paper and propose potential avenues for future research.

\section{Security Statement}
To prevent any potential misuse and ensure responsible use, we have strictly limited the application scope of our proposed method, \name. Unless authorized, \name\ is only permitted for use in research domains.

Additionally, access to the proposed dataset is restricted to qualified institutions and organizations, who must provide a clear purpose for its use. We explicitly prohibit the application of the dataset in situations that may cause potential danger or have a significant social impact. 

These measures are in place to ensure the ethical and responsible use of our research.

\section{Details in Training and Testing}
\paragraph{Computational Resource} During the pre-training phase, we utilized four Nvidia GTX A6000 GPUs\footnote{\url{https://www.nvidia.com/en-us/design-visualization/rtx-a6000/}} to train on the WebVid dataset~\cite{Bain21} for approximately 120 hours. In the fine-tuning phase, we employed two Nvidia GTX A6000 GPUs to fine-tune on our proposed dataset for about 80 hours.

\paragraph{Efficiency} During testing, the average model response time for each round of conversation with \name\ is approximately 2ms. Additionally, considering the available graphics memory, the model can handle video clips of up to 32 frames. Therefore, it is necessary to extract different frames from longer videos.

\paragraph{Hyper-parameters} In the loss function, $\mathbf{t_0}$ is set to 1 for our main task, video-to-language, and $\mathbf{t_1}$ and $\mathbf{t_2}$ are set to 0.1, as two auxiliary tasks for balancing different loss values.

\section{Details in Dataset} \label{appdata}
\paragraph{Dataset Introduction and Statistics}  Our study utilizes seven video anomaly datasets, each encompassing different scenes. The detailed statistics and introduction of these datasets are as follows:

\begin{itemize}
    \item  UCF-Cirme~\cite{sultani2018real}: The UCF-Crime dataset comprises an extensive collection of 128 hours of video. It consists of 1,900 long and untrimmed real-world surveillance videos, featuring 13 distinct classes of realistic anomalies. These anomalies are carefully chosen due to their notable implications for public safety. 
    \item  ShanghaiTech~\cite{liu2018ano_pred}: 
    The ShanghaiTech Campus dataset comprises 13 scenes characterized by complex light conditions and varied camera angles. It encompasses 130 instances of abnormal events and encompasses over 270,000 training frames. Notably, this dataset includes annotations for both frame-level and pixel-level ground truth of abnormal events, providing comprehensive insight into anomaly detection and localization tasks.
    \item  CUHK Avenue~\cite{lu2013abnormal}: The CUHK Avenue Dataset comprises 16 training and 21 testing video clips designed for abnormal event detection. Captured within the CUHK campus avenue, these videos encompass a total of 30,652 frames, divided into 15,328 frames for training and 15,324 frames for testing. The training videos capture normal situations, while the testing videos include both normal and abnormal events.
    \item UCSD Dataset~\cite{chan2008modeling,wang2010anomaly}: The UCSD Anomaly Detection Dataset was captured using a stationary camera positioned at an elevation, providing an overhead view of pedestrian walkways. The crowd density within these walkways exhibits variability, spanning from sparsely populated areas to densely crowded environments. It is split into 2 subsets, each corresponding to a different scene. Ped1~\cite{chan2008modeling} includes a total of 34 training video samples and 36 testing video samples, while Ped2~\cite{wang2010anomaly} consists of 16 training video samples and 12 testing video samples.
    \item  DoTA~\cite{yao2022dota}: The Detection of Traffic Anomaly (DOTA) Dataset introduces the When-Where-What pipeline with temporal, spatial, and categorical annotations. It contains 4677 videos, all with a resolution of 1280 x 720 pixels. Notably, the original videos were extracted at a frame rate of 10 fps in this dataset.
    \item  UBnormal~\cite{acsintoae2022ubnormal}: The UBnormal dataset is a supervised open-set benchmark designed explicitly for video anomaly detection, comprising diverse virtual scenes. It introduces abnormal events annotated at the pixel level during training, which enables the utilization of fully-supervised learning techniques for abnormal event detection.
\end{itemize}
In our study, we extend upon these existing datasets by implementing our data engineering pipeline. This pipeline generates comprehensive descriptions of video anomalies and formulates open questions derived from these anomalies.

\begin{figure}[h]
    \centering
    \includegraphics[width=0.8\linewidth]{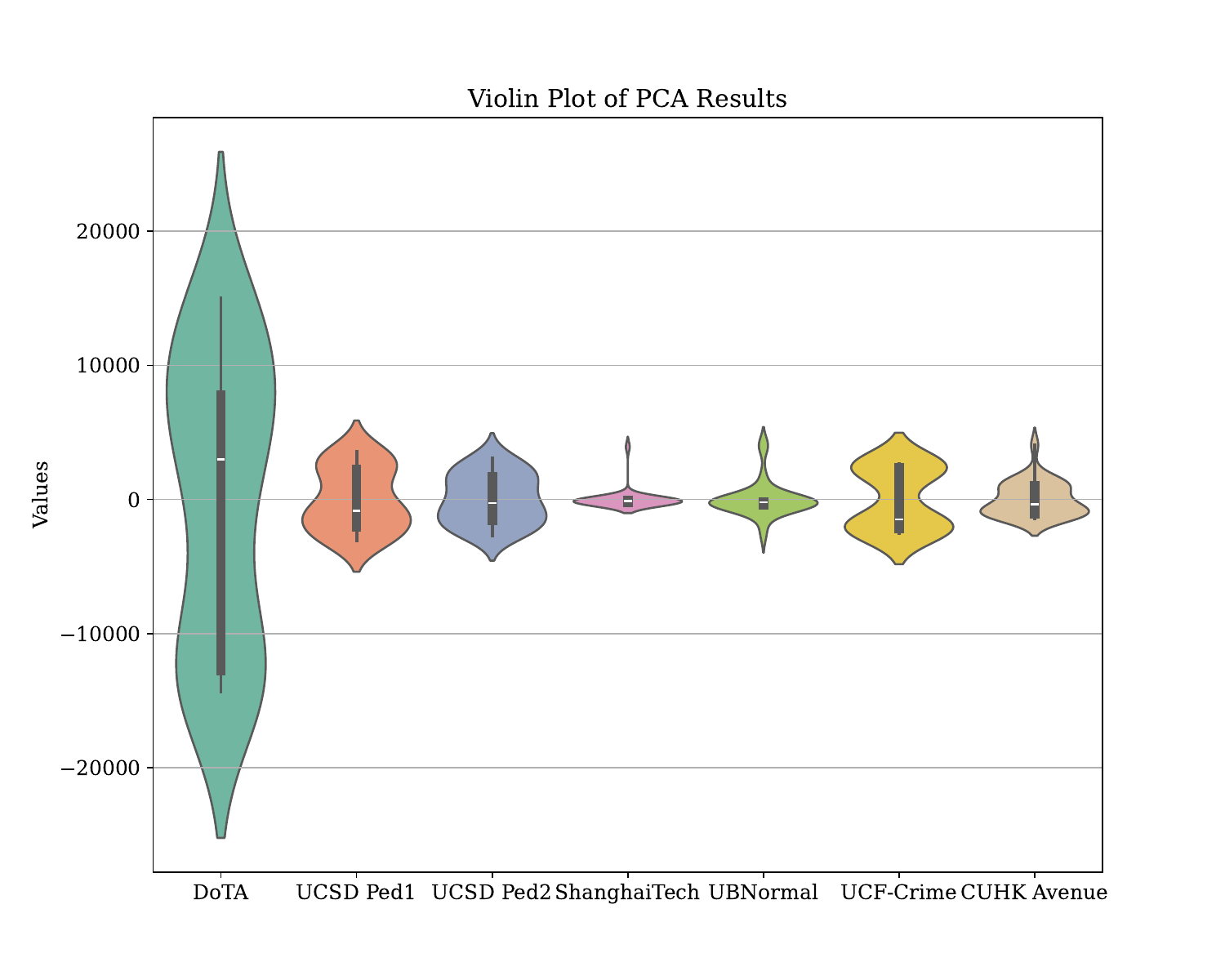}
    \caption{Violin plot of data distribution. We use PCA dimensional reduction to measure the feature distribution of different datasets, where there are significant differences in the feature distribution.}
    \label{distribution}
\end{figure}

\paragraph{Data Distribution}
To demonstrate the applicability of our data in an open-world scenario, we conducted a statistical analysis of the data distribution. Figure~\ref{distribution} illustrates the data distribution of all the datasets we utilized, indicating that our method can effectively support various open-world datasets. Besides, we acknowledge the need to expand our dataset further to enhance the model's applicability in this task.

\paragraph{Manual Checking} 
Before conducting the experiments, we performed the manual checking on the textual descriptions generated for the videos. Specifically, we consider the following aspects:
\begin{enumerate}
    \item Error Correction: We removed text descriptions that contained obvious errors about the video content and supplemented the correct object, behavior, and scene information. (For instance, GPT tends to misidentify dogs in videos, describe running pedestrians as skateboards and motorcycles, and mistake scenes containing water as rainy days.)
    \item Detail Enhancement: We provided more detailed textual descriptions of anomalies in the video (such as pedestrians lingering or jumping in the middle of the road).
    \item Human Resource Cost: We formed a team of five annotators to conduct Manual Checking on all the videos. Since most of the videos already had automatically generated annotations, each annotator invested approximately 30 hours of work during the labeling process, processing about 1700 videos.
\end{enumerate}

\paragraph{\textsc{<Describe\_Video>} and Generated Open-World \textsc{<Question>}} \label{prompt}
We set 20 problems for \textsc{<Describe\_Video>}, and during each iteration in training, we randomly select one of them.
\begin{scriptsize}\begin{minted}[bgcolor=lightgray!20, breaklines=true]{text}
1.	Can you describe the anomaly in the video?
2.	How would you detail the anomaly found in the video?
3.	What anomaly can you identify in the video?
4.	Could you explain the anomaly observed in the video?
5.	Can you point out the anomaly in the video?
6.	What's the anomaly depicted in the video?
7.	Could you specify the anomaly present in the video?
8.	How do you perceive the anomaly in the video?
9.	Can you highlight the anomaly within the video?
10.	What anomaly is noticeable in the video?
11.	Could you characterize the anomaly seen in the video?
12.	Can you detail the specific anomaly encountered in the video?
13.	How would you describe the particular anomaly in the video?
14.	What details can you provide about the anomaly in the video?
15.	Could you elucidate on the anomaly detected in the video?
16.	Can you illustrate the nature of the anomaly in the video?
17.	What features of the anomaly in the video can you describe?
18.	Could you outline the anomaly observed in the video?
19.	How does the anomaly in the video manifest?
20.	Can you clarify the aspects of the anomaly in the video?
\end{minted} 
\end{scriptsize}

We have also generated 100 \textsc{<Questions>} for open-world anomalies. To mimic user behavior, some of these questions are closely related to the video scene, while others are less closely related. However, all of these questions are potential inquiries in an open-world scenario.
\begin{scriptsize}\begin{minted}[bgcolor=lightgray!20, breaklines=true]{text}
1.	Who is causing the disturbance in the video?
2.	What is the unusual activity happening in the video?
3.	When did the anomaly occur in the video?
4.	Where is the strange event taking place in the video?
5.	Why is the object in the video behaving abnormally?
6.	How is the anomaly in the video affecting the surroundings?
7.	How much damage was caused by the incident in the video?
8.	Who is the main person involved in the unusual event?
9.	What is the cause of the sudden change in the video?
10.	When does the suspicious activity start in the video?
11.	Where can I find more information about the incident in the video?
12.	Why are the people in the video reacting in that way?
13.	How can I identify the source of the problem in the video?
14.	How much time does the abnormal event last in the video?
15.	Who are the other people affected by the anomaly in the video?
16.	What actions were taken to address the issue in the video?
17.	When was the video recorded, and is it a recent event?
18.	Where else can I find similar incidents in other videos?
19.	Why is the vehicle in the video moving erratically?
20.	How can I prevent such anomalies from occurring in the future?
21.	How much impact does the abnormal event have on the overall situation?
22.	Who should I contact if I notice a similar anomaly in another video?
23.	What steps can I take to investigate the issue further?
24.	When is the best time to report an unusual event in a video?
25.	Where can I find resources to help me understand the anomaly better?
26.	Why did the equipment in the video malfunction?
27.	How can I differentiate between normal and abnormal behavior in a video?
28.	How much does it cost to implement a system that detects anomalies in videos?
29.	Who can provide expert advice on handling video anomalies?
30.	What is the most common type of anomaly found in videos?
31.	When should I be concerned about an anomaly in a video?
32.	Where can I find a list of known video anomalies and their descriptions?
33.	Why is it important to detect and analyze anomalies in videos?
34.	How can I improve my ability to spot anomalies in videos?
35.	How much training is required to become proficient in detecting video anomalies?
36.	Who can I collaborate with to better understand video anomalies?
37.	What are the potential consequences of ignoring an anomaly in a video?
38.	When did the trend of analyzing anomalies in videos begin?
39.	Where can I find examples of successfully resolved video anomaly cases?
40.	Why do some anomalies in videos go unnoticed?
41.	How can I report a video anomaly to the appropriate authorities?
42.	How much time is needed to thoroughly analyze a video anomaly?
43.	Who is responsible for monitoring and addressing video anomalies?
44.	What are the best tools to use for detecting anomalies in videos?
45.	When is it necessary to escalate a video anomaly for further investigation?
46.	Where can I find guidelines on how to handle video anomalies?
47.	Why do some video anomalies lead to serious consequences?
48.	How can I ensure the accuracy of my video anomaly detection system?
49.	How much effort is needed to maintain a video anomaly detection system?
50.	Who should be informed when a video anomaly is detected?
51.	What are the signs that indicate a potential anomaly in a video?
52.	When should I perform a follow-up analysis on a detected video anomaly?
53.	Where can I find support for dealing with video anomalies?
54.	Why is it crucial to act quickly when a video anomaly is detected?
55.	How can I improve the efficiency of my video anomaly detection process?
56.	How much data is needed to accurately detect anomalies in videos?
57.	Who can help me fine-tune my video anomaly detection system?
58.	What are the key factors to consider when analyzing video anomalies?
59.	When should I update my video anomaly detection system?
60.	Where can I find the latest research on video anomaly detection techniques?
61.	Why is it necessary to have a video anomaly detection system in place?
62.	How can I minimize false alarms in my video anomaly detection system?
63.	How much does it cost to maintain a video anomaly detection system?
64.	Who can I consult if I encounter difficulties with my video anomaly detection system?
65.	What are the best practices for dealing with video anomalies?
66.	When is it appropriate to involve law enforcement in a video anomaly case?
67.	Where can I find a community of professionals who specialize in video anomaly detection?
68.	Why do some video anomalies require immediate attention?
69.	How can I enhance the performance of my video anomaly detection system?
70.	How much should I invest in a video anomaly detection system?
71.	Who can provide training on how to detect and analyze video anomalies?
72.	What are the most effective methods for detecting anomalies in videos?
73.	When should I seek external help for a video anomaly case?
74.	Where can I find a comprehensive database of video anomalies?
75.	Why is it important to continuously monitor videos for anomalies?
76.	How can I validate the results of my video anomaly detection system?
77.	How much influence do external factors have on video anomalies?
78.	Who can I reach out to for assistance with a complex video anomaly case?
79.	What are the main challenges in detecting and analyzing video anomalies?
80.	When is it necessary to involve other stakeholders in a video anomaly case?
81.	Where can I find case studies on successful video anomaly detection projects?
82.	Why is it essential to have a systematic approach to video anomaly detection?
83.	How can I optimize my video anomaly detection system for different scenarios?
84.	How much storage is needed to archive video anomalies for future analysis?
85.	Who should be held accountable for undetected video anomalies?
86.	What are the most common reasons for video anomalies to occur?
87.	When should I reevaluate my video anomaly detection system?
88.	Where can I find information on the latest video anomaly detection technologies?
89.	Why is it beneficial to collaborate with others in the field of video anomaly detection?
90.	How can I ensure the confidentiality of video anomaly cases?
91.	How much should I rely on automated systems for video anomaly detection?
92.	Who can I contact for technical support with my video anomaly detection system?
93.	What are the ethical considerations when dealing with video anomalies?
94.	When should I notify the public about a video anomaly case?
95.	Where can I find reliable sources of information on video anomalies?
96.	Why is it important to have a backup plan for dealing with video anomalies?
97.	How can I customize my video anomaly detection system for specific use cases?
98.	How much time should I allocate for analyzing video anomalies?
99.	Who can I turn to for guidance on handling sensitive video anomaly cases?
100.	What are the most critical factors to consider when choosing a video anomaly detection system?
\end{minted} 
\end{scriptsize}

\section{Details in GPT-Guided Metrics}
In the GPT-Guided metrics, we employ GPT-4 as an auxiliary tool to evaluate the generated response of \name. Our evaluation focuses on three primary dimensions: Reasonability, Detail, and Consistency. 

We first set the system prompt as follows: Initially, we establish the system prompt as shown below:
\begin{minted}[bgcolor=lightgray!20, breaklines=true]{json}
{"role": "system", "content": 
"You are an intelligent chatbot designed for evaluating the generative outputs for video-based pairs. you will be given two answers, one reference ground truth and one our generated, but this does not mean that the reference GT is the only answer. Your task is to give the score of the predicted answers."} 
\end{minted}
Our system prompt is designed to compare the degree of matching between image pairs. However, this does not imply fine-grained matching at the text level. Instead, it emphasizes the semantic information-related aspects.

To assess a particular dimension of the metric, we employ the following prompt: 
\begin{minted}[bgcolor=lightgray!20, breaklines=true]{json}
{"role": "user", "content":
"### Video Description Generation
Please evaluate the following video-based video description pair:
Reference: <DESCRIPTION_GT>
Ours: <DESCRIPTION_Ours>

### Video Question-Answering
Please evaluate the following video-based video question-answer pair:
Question: <QUESTION>
Reference: <ANSWER_GT>
Ours: <ANSWER_Ours>

Provide your evaluation only as a <Reasonability|Detail|Consistency> score where the <Reasonability|Detail|Consistency> score is a FLOAT value between 0 and 1, with 1 indicating the highest level of <Reasonability|Detail|Consistency>. Please generate the response in the form of a Python dictionary string with key 'score', where its value is the <Reasonability|Detail|Consistency> score in FLOAT, not STRING. DO NOT PROVIDE ANY OTHER OUTPUT TEXT OR EXPLANATION. Only provide the Python dictionary string. For example, your response should look like this: {'score': 0.675}."}
\end{minted}
We have developed distinct prompts for two tasks: Video Description Generation and Video Question-Answering. The primary difference is the addition of the <QUESTION> field in Video Question-Answering, which indicates what kind of question the model should answer. <DESCRIPTION\_GT> and <DESCRIPTION\_Ours> represent the Ground Truth and our generated video description, respectively. Similarly, <ANSWER\_GT> and <ANSWER\_Ours> signify the Ground Truth and our generated video answers, respectively. < Reasonability | Detail | Consistency > represents the three dimensions we aim to evaluate. Lastly, besides the essential reminders, we have constrained GPT's output format to \{`score': 0.675\}.

\section{More Results}

Table (A), (B), (C), (D), (E), and (F) below present additional qualitative results from different datasets. In the tables, \textcolor{red}{red} texts indicate key semantic inconsistencies with the Ground Truth, while \textcolor{ForestGreen}{green} texts signify that the generated results closely align with the Ground Truth.

 \section{Open-World Video Anomaly Understanding Demo}

\begin{figure}[p]
    \centering
    \includegraphics[width=1\linewidth]{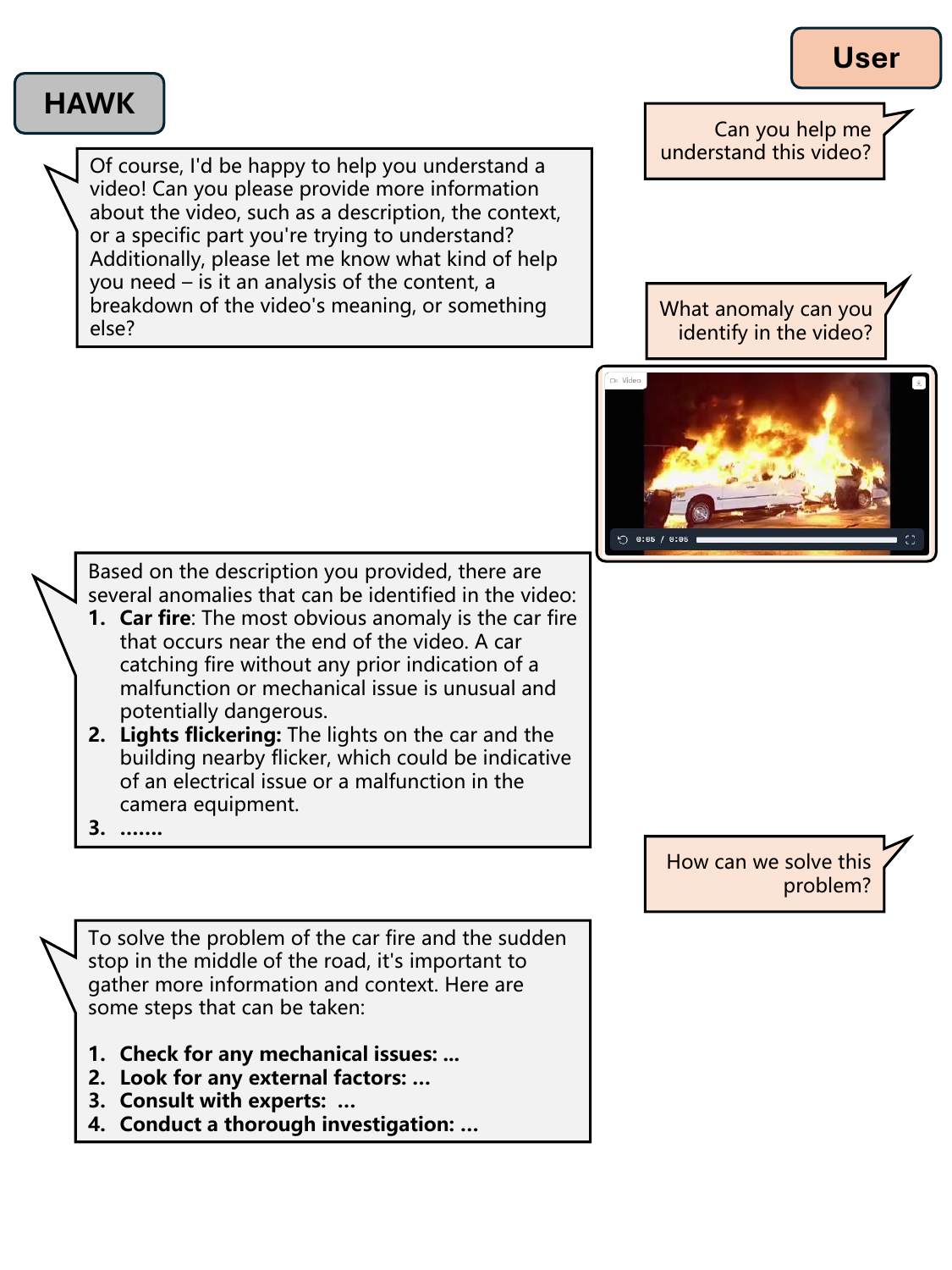}
    \caption{Demo on Open-World Video Anomaly Understanding. (Part 1)}
    \label{demo1}
\end{figure}

\begin{figure}
    \centering
    \includegraphics[width=1\linewidth]{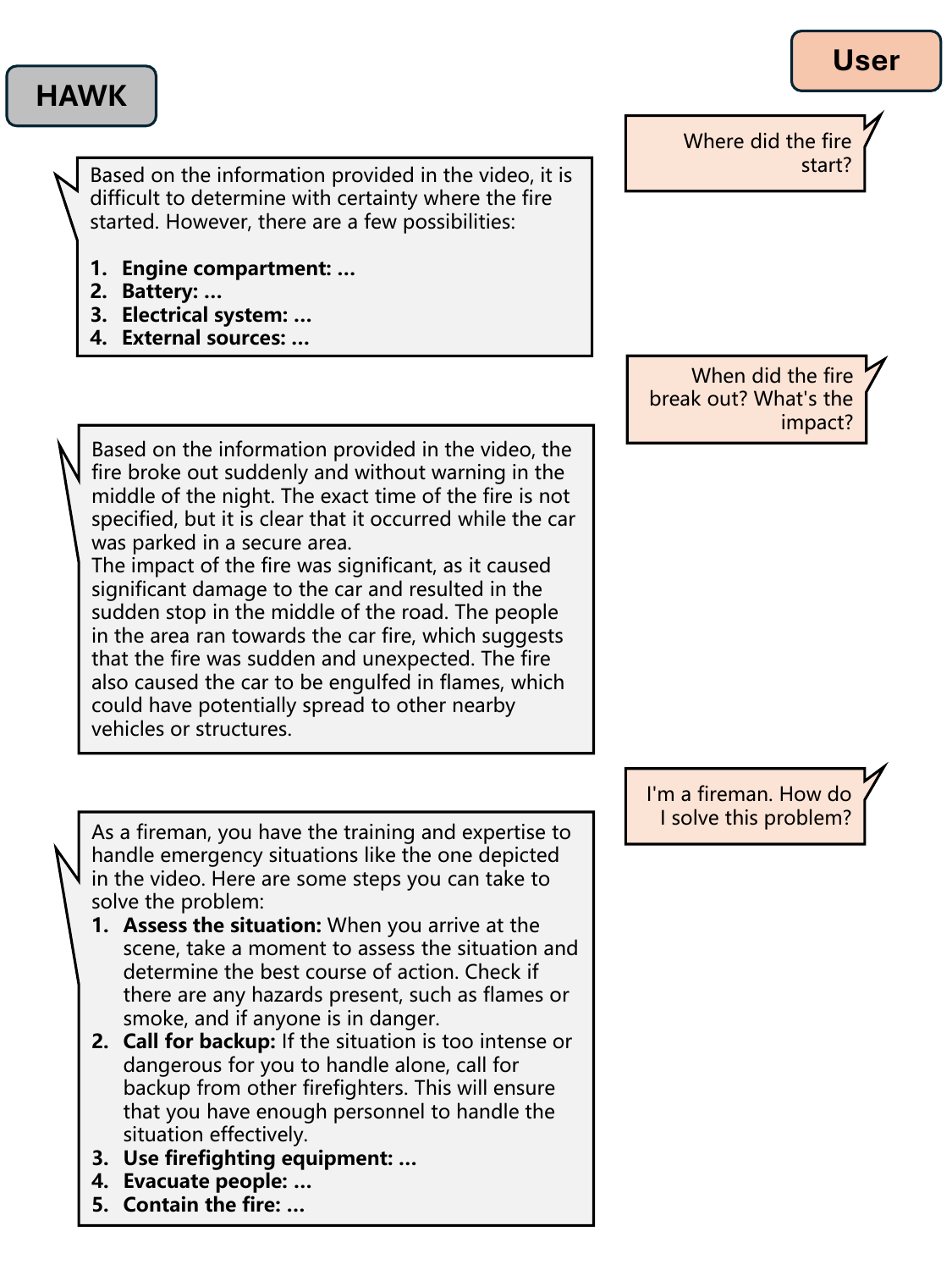}
    \caption{Demo on Open-World Video Anomaly Understanding. (Part 2)}
    \label{demo2}
\end{figure}

We present a demo showcasing the use of \name\ in an open-world scenario, using XD-Violence~\cite{Wu2020not} (which is not included in our dataset). The practical capability of the system in an unknown scenario in the open world is depicted in Fig.\ref{demo1} and Fig.\ref{demo2}. Furthermore, \name\ can provide accurate answers to users' questions and engage in long dialogues in the open world.

\section{Limitations}

\paragraph{Hallucination} Although most of the hallucinations can be decreased through motion, some error motion may still also cause hallucinations. Future work may need to consider the connection between the hallucination and the abnormal region more precisely.

\paragraph{Video-level v.s. Streaming Data} The goal of this paper is video-level video anomaly understanding. However, for a video anomaly detection system, anomaly detection in streaming is essential, so to increase the practical application ability, we need to design a more practical system for streaming data.

\paragraph{Data Limitations} While our dataset includes multiple anomaly scenarios and our framework is designed for an open-world setting, the limitations of our data make it difficult to fully support open-world scenarios. This is a significant drawback of our study. To address this limitation, we recommend building larger and more diverse open datasets.

\section{Future Work}
In this section, we discuss potential avenues for future research to build upon this paper.

\paragraph{Expanding Applicability} One possible direction for future work is to expand the applicability of the model to a wider range of scenarios (like most recent research by Du et al.~\cite{du2024uncovering}). While our current dataset includes multiple anomaly scenarios, there are still many diverse scenarios the model may not handle. We could explore ways to incorporate additional data sources or create new datasets that cover a broader range of scenarios. Additionally, we could investigate ways to deploy the model in cloud, edge, and hybrid computing environments to support more diverse deployment scenarios.

\paragraph{Enriching Task Capabilities} While our current model is able to detect anomalies in videos, it does not provide location information where the anomaly occurs. This information could be valuable for certain applications, such as security monitoring or surveillance. Therefore, a valuable direction for future work would be to investigate ways to improve the model's ability to localize anomalies. By improving the model's ability to locate the scene of an anomaly, we could further enhance its usability and expand its range of applications.


\newpage
\begin{tcolorbox}[title=(A) Anomaly Video Description Generation in DoTA~\cite{yao2022dota}.,
    colback=white, 
    colframe=black, 
    colbacktitle=black, 
    coltitle=white,
    fonttitle=\small\bfseries,
    enhanced,
    boxed title style={size=small,colback=green!20!white}] 
    \centering
    \small
    \includegraphics[width=1\linewidth]{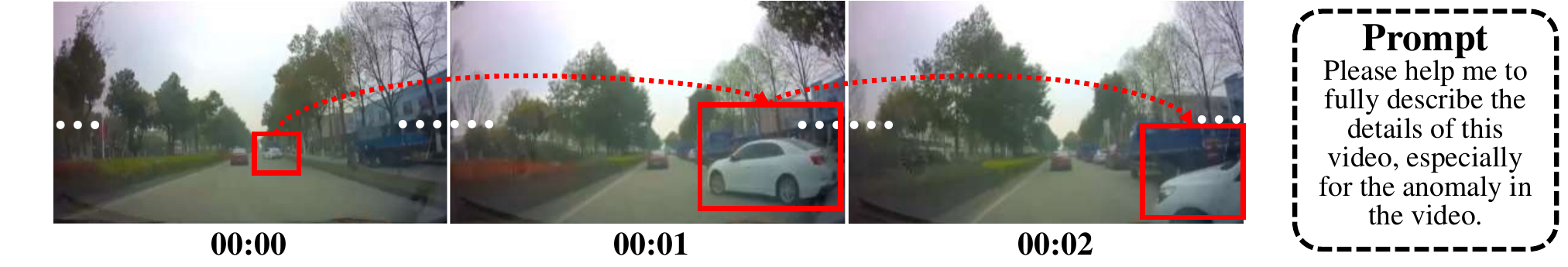} 
    \\ \rule[0\baselineskip]{\textwidth}{1pt}
    \begin{tabularx}{\textwidth}{@{}lX@{}}
    \textbf{Video-ChatGPT~\cite{Maaz2023VideoChatGPT}:} & The video shows a car driving down a busy \textcolor{ForestGreen}{street}, \textcolor{red}{with a person sitting in the passenger seat}. The car is surrounded by other cars, and the driver is seen looking at the camera. The car is seen turning a corner, and the person sitting in the passenger seat is looking at the camera. The video also shows \textcolor{red}{a person walking in the street, and a truck driving by.} \\
    \midrule
    \textbf{VideoChat~\cite{li2023videochat}:} & The video is a car driving on the road with the camera mounted inside the car. It is a clear day with some clouds in the sky. There are cars and other vehicles on the road, with one vehicle parked on the side of the road. The road is lined with trees on both sides. The car is driving along the road, passing other vehicles, and crossing roads. The anomaly in the video occurs when \textcolor{red}{the car suddenly stops in front of a bus, blocking the road.}\\
    \midrule
    \textbf{Video-LLaMA~\cite{damonlpsg2023videollama}:} & The car is driving down the \textcolor{ForestGreen}{street}, and there are buildings and cars parked along the side of the road. Finally, the car is seen driving down the street at night. The camera is shown outside the car, and there are white lights on the road. There are trees and buildings in the distance, and the cars on the road are all different colors. The video ends with \textcolor{red}{the car turning a corner and driving down the street at night.}\\
    \midrule
    \textbf{LLaMA-Adapter~\cite{zhang2023llamaadapter}:} & The video shows a car driving down a road, with a person inside the vehicle. Suddenly, the car appears to be \textcolor{red}{hit by a large object}, causing it to \textcolor{red}{flip over and crash}. The person inside the car is thrown out, and the car is left in a wrecked state on the road. The scene is captured from a dash cam perspective, providing a clear view of the incident. \\
    \midrule
    \textbf{Video-LLaVA~\cite{lin2023video}:} & The video starts with a car driving down a  \textcolor{ForestGreen}{street}, and the camera captures the view from inside the car. The car is seen driving past a \textcolor{red}{blue truck and a red car}. The camera then pans to the side of the road, where \textcolor{red}{a man} is seen walking down the sidewalk. The camera then pans back to the street, where a blue truck is seen driving down the road.\\
    \midrule
    \textbf{Ours:} & While driving down the \textcolor{ForestGreen}{street}, the current vehicle is involved in an accident when a \textcolor{ForestGreen}{sudden stop} causes it to rear-end \textcolor{ForestGreen}{the white car in front of it}, resulting in \textcolor{ForestGreen}{minor damage} to both vehicles and a startled reaction from the occupants. \\
    \midrule
    \textbf{Ground Truth:} & As the car continues down the busy tree-lined \textcolor{ForestGreen}{street}, the traffic light ahead suddenly turns red. \textcolor{ForestGreen}{The white car in front of the vehicle abruptly comes to a stop}, causing the driver recording the video to brake hard but ultimately \textcolor{ForestGreen}{colliding into the rear of the white car} with a loud impact. \\
    \end{tabularx}
     \\ \rule[0\baselineskip]{\textwidth}{1pt}
\end{tcolorbox}
\newpage
\begin{tcolorbox}[title=(B) Anomaly Video Description Generation in CUHK Avenue~\cite{lu2013abnormal}.,
    colback=white, 
    colframe=black, 
    colbacktitle=black, 
    coltitle=white,
    fonttitle=\small\bfseries,
    enhanced,
    boxed title style={size=small,colback=green!20!white}] 
    \small
    \centering
    \includegraphics[width=1\linewidth]{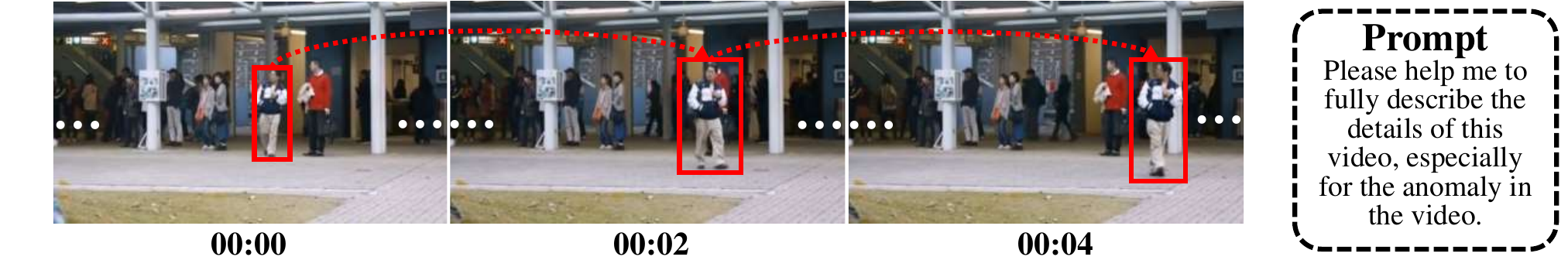} 
    \\ \rule[0\baselineskip]{\textwidth}{1pt}
    \begin{tabularx}{\textwidth}{@{}lX@{}}
    \textbf{Video-ChatGPT~\cite{Maaz2023VideoChatGPT}:} & The video shows a man sitting on a bench in a busy city street, surrounded by many people. Suddenly, a dog appears and starts walking towards the man. \textcolor{red}{The dog is wearing a red collar and is accompanied by a person. The dog is walking towards the man}, who is sitting on the bench, and the man is watching the dog as it approaches. The dog is walking towards the man, and it appears to be a unique and unexpected event in the busy city street. \\
    \midrule
    \textbf{VideoChat~\cite{li2023videochat}:} & The video shows a group of students walking outside the entrance to an office building located in a park. There are two white benches in front of the entrance, one on the left side and one on the right side. The entrance has \textcolor{red}{a large blue sign above it that reads Science and Engineering}. People are standing and walking around the entrance while the students are walking towards the building. \\
    \midrule
    \textbf{Video-LLaMA~\cite{damonlpsg2023videollama}:} & First, the video shows a busy street with many people walking around. There is a sign for a bus stop at one corner. Next, the video focuses on a line of people standing in front of a blue building, with a man in a black shirt and black jacket walking towards them. Finally, the video shows a crowd of people at the bus stop, with \textcolor{red}{people walking to and from the bus}. There is a green park bench and a building visible behind the people. The street is also filled with trees and \textcolor{red}{a black car} is visible in the background. \\
    \midrule
    \textbf{LLaMA-Adapter~\cite{zhang2023llamaadapter}:} & The video shows a group of people \textcolor{red}{walking down a hallway}, with some of them carrying backpacks. There are at least 12 people in the scene, with some walking in pairs or small groups. The hallway is long and has a few benches along the way. One of the people in the group appears to be wearing a tie, which stands out as an anomaly in the otherwise casual attire of the other individuals. \\
    \midrule
    \textbf{Video-LLaVA~\cite{lin2023video}:} & In the video, a man is seen standing in a hallway while \textcolor{red}{talking on his cell phone}. He is wearing a red shirt and is standing in front of a group of people who are walking by. The man's red shirt stands out against the crowd, making him easily noticeable. The man's behavior is quite unusual, as he is not paying attention to his surroundings while \textcolor{red}{talking on the phone}. It is unclear what the man is doing or why he is not paying attention to the people around him. \\
    \midrule
    \textbf{Ours:} & The video is set at a \textcolor{ForestGreen}{subway entrance} near campus, with \textcolor{ForestGreen}{pedestrians} entering and exiting the area from different directions. A crowd of students, teachers, and tourists is standing at the subway entrance, \textcolor{ForestGreen}{watching a performance} by a street musician. They are throwing coins and applauding. Other students and teachers are walking and talking near the subway entrance.  \\
    \midrule
    \textbf{Ground Truth:} & The video is set at a \textcolor{ForestGreen}{subway entrance} where some \textcolor{ForestGreen}{pedestrians} are hurriedly entering and exiting the subway or passing through nearby pathways. Another group of people is \textcolor{ForestGreen}{captivated by a performance} and stands at the subway entrance to watch. \\
    \end{tabularx}
     \\ \rule[0\baselineskip]{\textwidth}{1pt}
\end{tcolorbox}
\newpage
\begin{tcolorbox}[title=(C) Anomaly Video Description Generation in UCF-Crime~\cite{sultani2018real}.,
    colback=white, 
    colframe=black, 
    colbacktitle=black, 
    coltitle=white,
    fonttitle=\small\bfseries,
    enhanced,
    boxed title style={size=small,colback=green!20!white}] 
    \centering
    \small
    \includegraphics[width=1\linewidth]{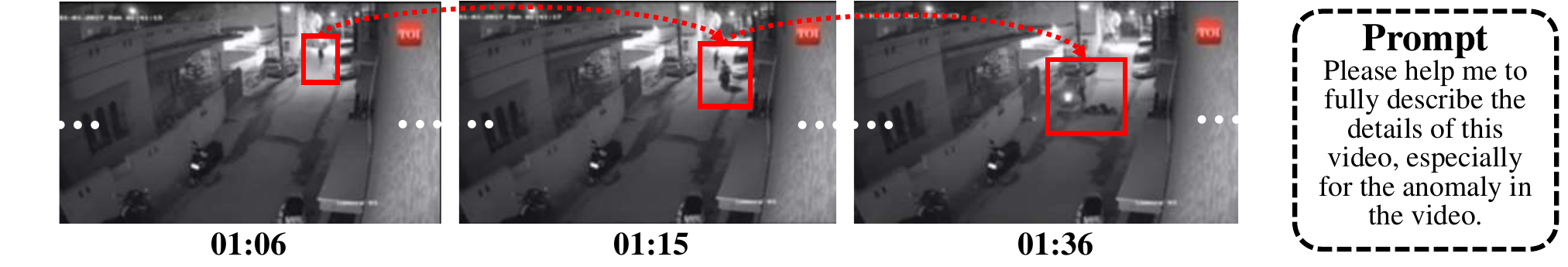} 
    \\ \rule[0\baselineskip]{\textwidth}{1pt}
    \begin{tabularx}{\textwidth}{@{}lX@{}}
    \textbf{Video-ChatGPT~\cite{Maaz2023VideoChatGPT}:} & The video shows a group of people riding motorcycles and a car in a city street. The video also shows a man walking in the street, and a man riding a motorcycle. The video is taken at night, and the street is dimly lit. The anomaly in the video is a person riding a motorcycle who is \textcolor{red}{not wearing a helmet}. This is a safety concern as it is illegal and can lead to severe injuries in case of an accident. \\
    \midrule
    \textbf{VideoChat~\cite{li2023videochat}:} & This video appears to be a surveillance video of a residential street at night in India. The video shows a street with some motorcycles parked along the side of the road and a few people walking around. However, \textcolor{red}{the two anomalies of the man riding the motorcycle and the man walking down the street make it stand out.}\\
    \midrule
    \textbf{Video-LLaMA~\cite{damonlpsg2023videollama}:} &First, the video shows a street in an Indian city at night. There are a few buildings on the street, and people are walking around. Next, there is a close-up of a street at night with cars parked on it. The street is narrow, and there are many buildings on both sides. Then, the video shows a street at night with a motorcycle parked on it. There is also \textcolor{red}{a bicycle parked on the sidewalk and a white car parked} next to a bike. \\
    \midrule
    \textbf{LLaMA-Adapter~\cite{zhang2023llamaadapter}:} & The video shows a man walking down a street at night, with a car parked nearby. Suddenly, the scene shifts to a different location, where a woman is walking down the street. The man and the woman appear to be walking in opposite directions, and the man is carrying a backpack. The video ends with the man walking down the street, still carrying the backpack. \\
    \midrule
    \textbf{Video-LLaVA~\cite{lin2023video}:} & The video shows a person riding a motorcycle down a narrow alley at night. The person is wearing a white shirt and black pants. The person rides the motorcycle quickly, and the camera follows them until they reach the end of the alley. However, there is an anomaly in the video. \textcolor{red}{The person on the motorcycle is not visible in the video, and the camera only shows the motorcycle and the surrounding environment.} \\
    \midrule
    \textbf{Ours:} & The video shows what appears to be a \textcolor{ForestGreen}{hit-and-run incident} or an attempted theft. A man in blue pants and shorts is shown walking along the side of the road, where he is subsequently struck by a motorcycle traveling from behind. The man, possibly injured, then \textcolor{ForestGreen}{falls to the ground}, which could be indicative of either the collision itself or a previous event that has left him unconscious but not necessarily injured.  \\
    \midrule
    \textbf{Ground Truth:} & The video portrays routine urban activity that people and vehicles moving along a street. The abnormal part unfolds when a motorcycle stops next to a pedestrian, leading to a confrontation. The altercation becomes physical when the person who dismounted from the motorcycle starts arguing with the passerby, eventually escalating as the passerby is \textcolor{ForestGreen}{pushed to the ground}. Following the altercation, the individuals involved in the aggressive act hastily escape the scene on the motorcycle, which adds a \textcolor{ForestGreen}{hit-and-run element to the incident}, indicating a consciousness of guilt or an attempt to evade law enforcement.  \\
    \end{tabularx}
     \\ \rule[0\baselineskip]{\textwidth}{1pt}
\end{tcolorbox}
\newpage
\begin{tcolorbox}[title=(D) Anomaly Video Question-Answering in ShanghaiTech~\cite{liu2018ano_pred}.,
    colback=white, 
    colframe=black, 
    colbacktitle=black, 
    coltitle=white,
    fonttitle=\small\bfseries,
    enhanced,
    boxed title style={size=small,colback=yellow!20!white}] 
    \small
    \includegraphics[width=\linewidth]{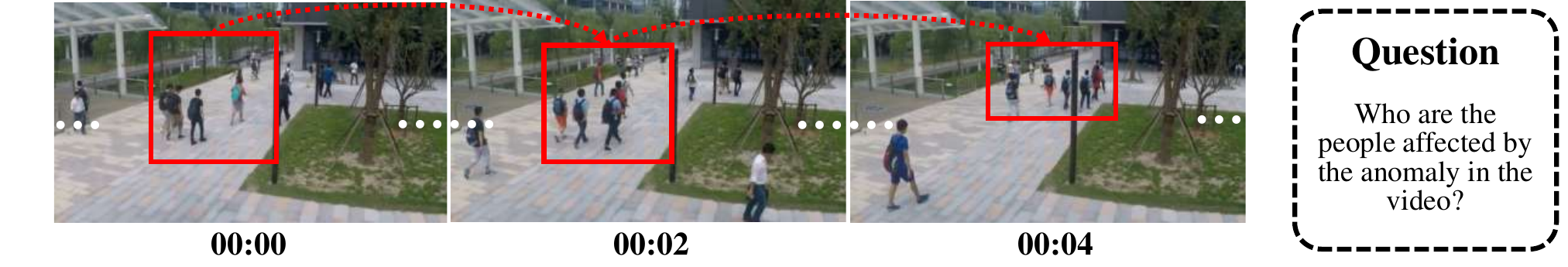} 
     \\ \rule[0\baselineskip]{\textwidth}{1pt}
    \begin{tabularx}{\textwidth}{@{}lX@{}}
    \textbf{Video-ChatGPT~\cite{Maaz2023VideoChatGPT}:} & The other people affected by the anomaly in the video are \textcolor{ForestGreen}{the people walking on the sidewalk}. They are walking in a line, and suddenly, a person starts running in the opposite direction. \textcolor{red}{The person running is wearing a red shirt.} \\
    \midrule
    \textbf{VideoChat~\cite{li2023videochat}:} & It is possible that there may be other people in the video who cross the yellow line, although they are not visible in the frame at that point in the video. \textcolor{red}{The pedestrian who crosses the traffic line} appears to be walking away from the camera and appears to be aware of the traffic line, which suggests that they are not necessarily \textcolor{red}{unaware of the rules of the road.} \\
    \midrule
    \textbf{Video-LLaMA~\cite{damonlpsg2023videollama}:} & It only talks about a group of people walking down the street. It does \textcolor{red}{not mention any specific people being affected by the anomaly.}\\
    \midrule
    \textbf{LLaMA-Adapter~\cite{zhang2023llamaadapter}:} & The other people affected by the anomaly in the video are the people walking on the sidewalk, who are \textcolor{red}{unaware of the man's disappearance.} \\
    \midrule
    \textbf{Video-LLaVA~\cite{lin2023video}:} & The video shows that the other people are also affected by the anomaly. They are seen \textcolor{red}{running and jumping over the obstacle} in the video.\\
    \midrule
    \textbf{Ours:} & The anomaly in the video that I described involves a man walking on \textcolor{ForestGreen}{a pedestrian walkway while other people are nearby, which could potentially cause a collision or accident}. Additionally,  \textcolor{ForestGreen}{other pedestrians or passersby in the vicinity may also be affected by the situation, potentially being startled or distracted} by the incident. Additionally, other pedestrians or passersby in the vicinity may also be affected by the situation, potentially being startled or distracted by the incident.\\
    \midrule
    \textbf{Ground Truth:} & The other people affected by the anomaly in the video would include: 1. The \textcolor{ForestGreen}{pedestrian who was collided with by the running man}, as they might sustain injuries from the impact. 2.  \textcolor{ForestGreen}{Bystanders or other individuals in the park who may witness} the incident and could potentially be involved in the ensuing chaos or need to provide assistance. 3. Any friends or family members of the individuals directly involved in the collision who might be present or affected indirectly by the consequences of the incident.\\
    \end{tabularx}
     \\ \rule[0\baselineskip]{\textwidth}{1pt}
\end{tcolorbox}
\newpage
\begin{tcolorbox}[title=(E) Anomaly Video Question-Answering in UBNormal~\cite{acsintoae2022ubnormal}.,
    colback=white, 
    colframe=black, 
    colbacktitle=black, 
    coltitle=white,
    fonttitle=\small\bfseries,
    enhanced,
    boxed title style={size=small,colback=yellow!20!white}] 
    \small
    \includegraphics[width=\linewidth]{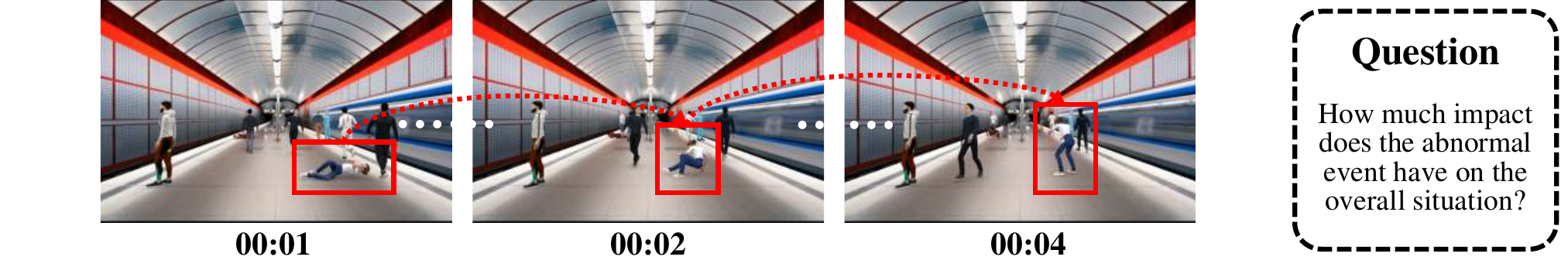} 
     \\ \rule[0\baselineskip]{\textwidth}{1pt}
    \begin{tabularx}{\textwidth}{@{}lX@{}}
    \textbf{Video-ChatGPT~\cite{Maaz2023VideoChatGPT}:} & The abnormal event of \textcolor{ForestGreen}{a person falling on the ground in the middle of the train station} has a significant impact on the overall situation. It disrupts the flow of people walking around the station, as they are forced to stop and look at the person lying on the ground. This creates a sense of confusion and concern among the people, as they may not know what to do or how to help the person. \\
    \midrule
    \textbf{VideoChat~\cite{li2023videochat}:} & Based on the information provided, \textcolor{red}{it's unclear if there is any abnormal event} occurring in the video. I cannot assess the overall situation without further context or information. \\
    \midrule
    \textbf{Video-LLaMA~\cite{damonlpsg2023videollama}:} & The abnormal event of \textcolor{ForestGreen}{a person lying on the ground} could disrupt the normal functioning of the metro station. It may cause delays or disruptions in the train service, potentially affecting passengers' plans and travel experiences. Additionally, first responders may have to intervene to ensure the person's safety.\\
    \midrule
    \textbf{LLaMA-Adapter~\cite{zhang2023llamaadapter}:} & The abnormal event of \textcolor{red}{a train passing through a tunnel with no passengers or train cars} significantly impacts the overall situation. In the image, the tunnel is empty, which is unusual as trains are typically filled with passengers or cargo. This lack of activity could be due to various reasons, such as maintenance work, a temporary suspension of services, or even an emergency situation.  \\
    \midrule
    \textbf{Video-LLaVA~\cite{lin2023video}:} & The abnormal event of \textcolor{ForestGreen}{the man falling down} has a significant impact on the overall situation. It disrupts the normal flow of the game and adds an element of surprise and unpredictability. The players' reactions to the event, such as the man getting up and continuing to play, show that the game continues despite the unexpected interruption.\\
    \midrule
    \textbf{Ours:} & Based on the description provided, the abnormal event in the video, which is the man in blue \textcolor{ForestGreen}{lying on the ground}, could potentially have a significant impact on the overall situation. This is because \textcolor{ForestGreen}{the man's unexpected appearance on the ground could cause a distraction or confusion among the other passengers}, which could in turn affect the smooth operation of the train and the evacuation process if necessary. Additionally, if the man is injured, it could also impact the response time and resources needed to address the situation.\\
    \midrule
    \textbf{Ground Truth:} & Based on the description provided, the abnormal event of \textcolor{ForestGreen}{a person collapsing or someone appearing intoxicated and losing their balance} has a significant impact on the overall situation. Such an event can lead to serious consequences, including severe injury or fatality, if it occurs near train tracks and coincides with the arrival of a train. The risk of a train not being able to stop in time and the individual's inability to move out of harm's way due to their condition magnifies the potential danger and impact of the event. Therefore, the impact is considered critical to \textcolor{ForestGreen}{the safety of the individuals involved and any potential bystanders.} \\
    \end{tabularx}
     \\ \rule[0\baselineskip]{\textwidth}{1pt}
\end{tcolorbox}
\newpage
\begin{tcolorbox}[title=(F) Anomaly Video Question-Answering in UCSD Ped1~\cite{chan2008modeling}.,
    colback=white, 
    colframe=black, 
    colbacktitle=black, 
    coltitle=white,
    fonttitle=\small\bfseries,
    enhanced,
    boxed title style={size=small,colback=yellow!20!white}] 
    \small
    \includegraphics[width=\linewidth]{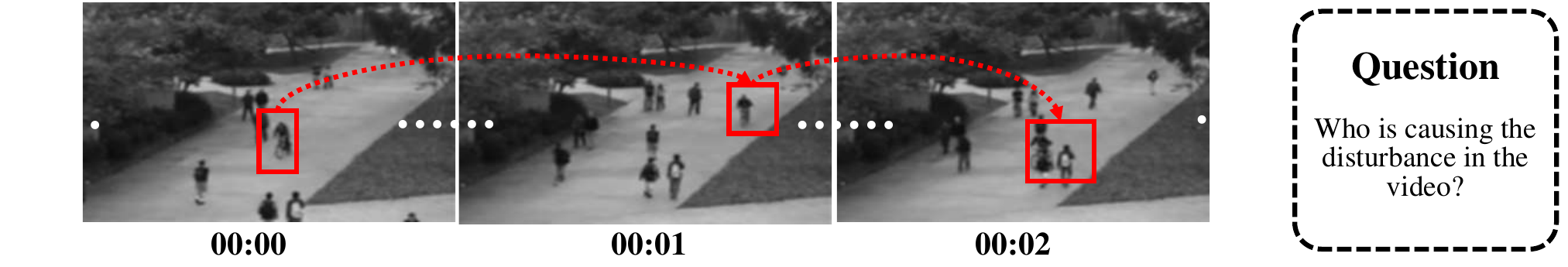} 
     \\ \rule[0\baselineskip]{\textwidth}{1pt}
    \begin{tabularx}{\textwidth}{@{}lX@{}}
    \textbf{Video-ChatGPT~\cite{Maaz2023VideoChatGPT}:} & The disturbance in the video is caused by \textcolor{ForestGreen}{a group of people walking down a sidewalk.} \\
    \midrule
    \textbf{VideoChat~\cite{li2023videochat}:} & There is no clear indication of any disturbance in the video. The students are simply walking down the sidewalk, and there are \textcolor{red}{no objects or actions that appear to be causing any disruption}. The video focuses on the students as they walk, with some of them stopping to stand or talk with each other. There are no sounds or other indications of any disturbance in the video. \\
    \midrule
    \textbf{Video-LLaMA~\cite{damonlpsg2023videollama}:} & There is \textcolor{red}{no disturbance} in the video. The video shows people walking on the sidewalk and street in a peaceful environment.\\
    \midrule
    \textbf{LLaMA-Adapter~\cite{zhang2023llamaadapter}:} & The disturbance in the video is caused by a group of people \textcolor{red}{walking and talking, creating a crowd.} \\
    \midrule
    \textbf{Video-LLaVA~\cite{lin2023video}:} & The video does not provide any information about the cause of the disturbance. However, it is clear that the people in the video are \textcolor{red}{trying to avoid the disturbance and continue on their way.}\\
    \midrule
    \textbf{Ours:} & Based on the description provided, \textcolor{ForestGreen}{the cyclist} is causing the disturbance in the video. The cyclist is riding on a path that is designated for pedestrians only, which is the anomaly that could lead to \textcolor{ForestGreen}{an accident and disturbance}. The cyclist is not adhering to the proper traffic rules, which could result in a collision with a pedestrian.\\
    \midrule
    \textbf{Ground Truth:} & Based on the description provided, it suggests that either \textcolor{ForestGreen}{the cyclist} or \textcolor{ForestGreen}{the pedestrian}, or both, could be causing the disturbance by being on \textcolor{ForestGreen}{a collision course}. Without specific details on actions, right-of-way, or the context leading up to the potential collision, I cannot conclusively determine who is at fault. \\
    \end{tabularx}
     \\ \rule[0\baselineskip]{\textwidth}{1pt}
\end{tcolorbox}


\end{document}